\documentclass[conference]{IEEEtran}
\IEEEoverridecommandlockouts
\usepackage{amsmath,amsfonts}
\usepackage{array}
\usepackage{textcomp}
\usepackage{stfloats}
\usepackage{url}
\usepackage{verbatim}
\usepackage{graphicx}
\usepackage{amssymb}
\usepackage{hyperref}
\usepackage{graphicx}
\usepackage{amsmath}
\usepackage{longtable}
\usepackage{algorithm} 
\usepackage{algpseudocode} 
\usepackage{mathrsfs}
\usepackage{subcaption}
\usepackage{mathtools}
\usepackage{pifont}
\usepackage{color}
\usepackage{lineno}
\usepackage{makecell} 
\usepackage{pdflscape}
\usepackage{adjustbox}
\usepackage[utf8]{inputenc}
\usepackage{tabularx}
\usepackage{blindtext}
\usepackage{longtable}
\usepackage{lscape}
\usepackage{setspace}
\usepackage{notoccite} 
\usepackage{lscape} 
\usepackage{mwe}
\usepackage{booktabs}
\usepackage{amsthm}

\theoremstyle{definition}

\theoremstyle{definition}

\newcommand{\RNum}[1]{\lowercase\expandafter{\romannumeral #1\relax}}
\newcommand{\RNumU}[1]{\uppercase\expandafter{\romannumeral #1\relax}}
\usepackage[numbers]{natbib}

\def\BibTeX{{\rm B\kern-.05em{\sc i\kern-.025em b}\kern-.08em
    T\kern-.1667em\lower.7ex\hbox{E}\kern-.125emX}}
\begin{document}

\title{Robust Universum Twin Support Vector Machine for Imbalanced Data\\
}

\author{
\IEEEauthorblockN{M. Tanveer}
\IEEEauthorblockA{
\textit{Department of Mathematics} \\
\textit{Indian Institute of Technology Indore}\\
mtanveer@iiti.ac.in}
\and
\IEEEauthorblockN{A. Quadir}
\IEEEauthorblockA{
\textit{Department of Mathematics} \\
\textit{Indian Institute of Technology Indore}\\
mscphd2207141002@iiti.ac.in}
}

\maketitle

\begin{abstract}
One of the major difficulties in machine learning methods is categorizing datasets that are imbalanced. This problem may lead to biased models, where the training process is dominated by the majority class, resulting in inadequate representation of the minority class. Universum twin support vector machine (UTSVM) produces a biased model towards the majority class, as a result, its performance on the minority class is often poor as it might be mistakenly classified as noise. Moreover, UTSVM is not proficient in handling datasets that contain outliers and noises. Inspired by the concept of incorporating prior information about the data and employing an intuitionistic fuzzy membership scheme, we propose intuitionistic fuzzy universum twin support vector machines for imbalanced data (IFUTSVM-ID) by enhancing overall robustness. We use an intuitionistic fuzzy membership scheme to mitigate the impact of noise and outliers.  Moreover, to tackle the problem of imbalanced class distribution, data oversampling and undersampling methods are utilized. Prior knowledge about the data is provided by universum data. This leads to better generalization performance. UTSVM is susceptible to overfitting risks due to the omission of the structural risk minimization (SRM) principle in their primal formulations. However, the proposed IFUTSVM-ID model incorporates the SRM principle through the incorporation of regularization terms, effectively addressing the issue of overfitting. We conduct a comprehensive evaluation of the proposed IFUTSVM-ID model on benchmark datasets from KEEL and compare it with existing baseline models. Furthermore, to assess the effectiveness of the proposed IFUTSVM-ID model in diagnosing Alzheimer's disease (AD), we applied them to the Alzheimer's Disease Neuroimaging Initiative (ADNI) dataset. Experimental results showcase the superiority of the proposed IFUTSVM-ID models compared to the baseline models. The supplementary material of the paper can be accessed using the following link: \url{https://github.com/AnonymousAuthor0011/IFUTSVM-ID.git}.
\end{abstract}


\section{Introduction}
\IEEEPARstart{S}{upport} vector machines (SVMs) \cite{cortes1995support} are advanced machine learning models that maximize the margin between two classes in a classification problem, aiming to find the optimal hyperplane between two parallel supporting hyperplanes. SVM has proven its remarkable utility across diverse domains such as fault detection \cite{zhang2017rotating}, text categorization \cite{joachims2005text}, and disease diagnosis \cite{tomar2015hybrid, richhariya2018eeg} and so on. SVM integrates the structural risk minimization (SRM) principle within its optimization framework, thereby enhancing its generalization capabilities by minimizing an upper bound of the generalization error. SVM solves one large quadratic programming problem (QPP), resulting in escalated computational complexity, which renders it less suitable for large-scale datasets. To enhance the efficiency of SVM, \citet{khemchandani2007twin} proposed twin support vector machine (TSVM). TSVM generates a pair of non-parallel hyperplanes, with each hyperplane deliberately situated in close proximity to the data points belonging to one class while ensuring a minimum separation distance of at least one unit from the data points of the other class. TSVM solves two smaller-sized QPPs instead of a single large QPP, making TSVM four times faster than the standard SVM. However, TSVM faced two notable challenges: the requirement for matrix inversions and the absence of the SRM principle in its formulation, which posed significant obstacles to its effectiveness. \citet{shao2011improvements} introduced twin-bounded SVM (TBSVM), which incorporated a regularization term in their formulation, allowing the principle of SRM to be employed. In recent years, numerous variants of TSVM have been proposed, including granular ball TSVM \cite{quadir2024granularGB, quadir2024granular}, robust energy-based LTSVM (RELTSVM) \cite{tanveer2016robust}, multiview learning with twin parametric margin SVM (MvTPMSVM) \cite{quadir2024multiview}, Wave-MvSVM \cite{quadir2024enhancing}, twin restricted kernel machine (TRKM) \cite{quadir2025trkm}, one class restricted kernel machine (OCRKM) \cite{quadir2025one} and others.

Several variants of the SVM model have been developed with the aim of improving its generalization performance; however, specific attributes of the original data may have been disregarded. In practical applications, it is common to encounter data that do not pertain to our target class but are within the same domain. These data, known as Universum data, can be utilized to enhance the learning process. Universum data consists of samples that do not belong to any of the relevant classes, providing insights into the distribution of the data. By integrating universum data into the SVM classifier, \citet{weston2006inference} proposed universum SVM (USVM). In USVM, the universum data points are integrated within the $\epsilon$-insensitive tube positioned between the binary classes. As a result, the resulting model demonstrates enhanced generalization performance. Universum twin support vector machine (UTSVM) \cite{qi2012twin} has been introduced to decrease the training time of USVM. Furthermore, many variants of SVM and TSVM based on universum data have been proposed over the years to enhance the efficiency of the models such as non-parallel hyperplane universum SVM (U-NHSVM) \cite{zhao2019improved}, lagrangian-based approach for universum twin-bounded SVM \cite{moosaei2023lagrangian}, multi-view UTSVM with Insensitive Pinball Loss (Pin-MvUTSVM) \cite{lou2024multi}, twin parametric margin SVM based on Universum data (UTPMSVM) \cite{hazarika2023eeg} and so on. 

While TSVM and its variants effectively handle the computational complexity challenges posed by SVM, both SVM and TSVM  encounter difficulties when dealing with class-imbalanced data, particularly in datasets containing noise and outliers. The accuracy of SVM and TSVM models' predictions is significantly affected by noisy data, outliers, and imbalanced datasets. The issue of class imbalance occurs when there is a notable difference in the number of samples belonging to one particular class compared to the others within a dataset. Although a uniform weighting scheme is applied to each data sample during the generation of the optimal classifier by SVM and TSVM, they are still susceptible to issues such as noise, outliers, and class imbalance. The fuzzy theory has proven to be effective in mitigating the adverse impacts of noise or outliers. Many fuzzy scheme-based approaches in SVM have been introduced so far for the classification of noisy data points \cite{quadir2024intuitionistic, akhtar2024flexi}. In classification problems where there is a significant imbalance between samples of majority and minority class,  SVM classification models often exhibit a biasness towards the majority class samples, which can lead to incorrect classification of samples of the minority class. To address the issue of class imbalance, an entropy-based approach proposed in \cite{fan2017entropy}. This approach assigns higher membership to each individual data point lying on the supporting hyperplane. \citet{richhariya2019fuzzy} employed a fuzzy membership scheme in both FUSVM and FUTSVM considering the probability of uncertainty of data points belonging or not belonging to the respective class. 
In the presence of noise and outliers, classification models encounter difficulties with imbalanced classes. 
In situations such as disease diagnosis \cite{richhariya2018eeg}, the primary objective is to accurately classify minority class samples. Various models are available to tackle the problem of imbalanced classes, such as fuzzy SVM (FSVM) for class imbalance learning \cite{li2013fuzzy}, which assigns fuzzy membership weights to the data samples. To address class imbalance learning, robust fuzzy LSTSVM (RFLSTSVM-CIL) \cite{richhariya2018robust} utilized the imbalance ratio to the data points to generate fuzzy weights, leading to an improvement in generalization performance. In order to mitigate the impact of noise, some approaches have been proposed for class imbalance problems in \cite{xu2019knn}. 
 TSVM remains sensitive to datasets that contain noise and outliers. In \cite{atanassov1999intuitionistic}, a variant of the fuzzy set known as the intuitionistic fuzzy set (IFS) is proposed. IFS incorporates both membership and non-membership functions, whereas a fuzzy set is only based on membership functions. This is the primary difference between IFS and a fuzzy set. In order to reduce the impact of noise and outliers, a combination of TSVM and IFS was proposed in \cite{rezvani2019intuitionistic}.

Prior knowledge has been shown to improve the effectiveness of universum-based learning algorithms, as demonstrated in \cite{weston2006inference,qi2012twin}. To address the imbalance in the classes, reduced UTSVM for class imbalance learning (RUTSVM-CIL) is introduced in \cite{richhariya2021efficient}. By Incorporating universum data into the model, prior knowledge about the data can be leveraged without significantly escalating the computational burden. We propose a novel intuitionistic fuzzy universum twin support vector machine for imbalanced data (IFUTSVM-ID). IFUTSVM-ID leverages the intuitionistic fuzzy (IF) theory, which utilizes membership and non-membership functions to assign an IF score to each sample, thereby efficiently managing noise and outliers and tackling the issue of imbalanced datasets in a comprehensive manner. The degree of membership function is calculated by the distance between the samples and the corresponding class center while the non-membership functions leverage the statistical correlation between the count of heterogeneous samples to all the samples within their neighborhoods. This membership scheme allows the model to effectively handle outliers and noise that have trespassed in the dataset. Furthermore, Universum data is integrated into the model to offer prior information that does not pertain to any specific target category but is within the same domain. The proposed IFUTSVM-ID model employs reduced kernel techniques to further simplify and reduce their complexity. The noteworthy contributions of the proposed IFUTSVM-ID model are outlined below:
\begin{itemize}
\item We propose an intuitionistic fuzzy universum twin support vector machine for
imbalanced data (IFUTSVM-ID). IFUTSVM-ID assigns  IF membership weight to each data point in the training dataset, with the aim of mitigating the adverse effect of noise and outliers.
\item The proposed IFUTSVM-ID applied the concept of undersampling to both the constraints and the kernel matrix. This ensures equal weighting of the two classes during classifier construction. The reduced kernel matrix is obtained by selecting a random subset of the original kernel matrix, leading to reduced computational complexity and storage requirements for the kernel matrices.
\item  The IFUTSVM-ID aims to reduce the structural risk inherent in its formulation by incorporating the SRM principle through the addition of a regularization term in its primal formulation. The proposed IFUTSVM-ID exhibits efficiency, scalability, proficient handling of overfitting, robustness, and improved generalization performance.
\item We performed experiments on $46$ real-world KEEL datasets. Numerical experiments and statistical analyses validate the superiority of the proposed IFUTSVM-ID model over the baseline models.
\item To demonstrate the resilience of the proposed IFUTSVM-ID model, even in challenging conditions, we deliberately introduced label noise into five varied datasets. Through experiments conducted in noisy environments, we provide compelling evidence showcasing the superiority of the proposed IFUTSVM-ID model.
\item As an application, the proposed IFUTSVM-ID model is employed for Alzheimer’s disease detection. Experimental results illustrate that the proposed IFUTSVM-ID model surpasses the baseline models in terms of accuracy. 
\end{itemize}

The remainder structure of this paper as follows: Section \ref{Related work} briefly explains the formulation of UTSVM and intuitionistic fuzzy membership scheme. In Section \ref{Intuitionistic Fuzzy Universum Twin Support Vector Machine for Imbalanced Data}, we derive the mathematical formulation of the proposed IFUTSVM-ID model. Numerical experiment comparisons are
demonstrated in Section \ref{Experimental results}. Conclusions and future directions are presented in section \ref{Conclusion}.

\section{Related works}
\label{Related work}
 In this section, we go through the mathematical formulation of UTSVM. The Intuitionistic fuzzy membership Scheme is discussed in Section S.I of the Supplementary Material.

\subsection{Universum Twin Support Vector Machine (UTSVM)}

Let $X_1$ and $X_2$ are two matrices that represent data samples assigned to positive and negative classes of dimensions $m_{1} \times n$ and $m_{2} \times n$, respectively, where $m_{1}$ ($m_{2}$) is the total number of samples in positive class (negative class) and $n$ represents the number of features in each sample. The universum data matrix $U$ has dimensions of $u \times n$.

UTSVM improved generalization by incorporating universum data, thus utilizing prior knowledge embedded in the universum dataset. The formulation of UTSVM can be expressed as follows:
{
\begin{align}
\label{eq:9}
\underset{ w_{1}, b_{1}, \xi_{1},\psi_{1}}{min}  \hspace{0.5cm}~&\frac{1}{2}\|X_1 w_{1} + e_{1} b_{1}\|^2+c_{1} e_{2}^t \xi_{1} + c_{u}e_{u}^t \psi_{1} \nonumber \\
 \text { s.t. }\hspace{0.5cm}  & -\left(X_2 w_{1} +e_{2}b_{1}\right) \geq e_{2}-\xi_{1}, \nonumber \\
 & \left(U w_{1} +e_{u}b_{1}\right) \geq (-1+\epsilon)e_{u}-\psi_{1}, \nonumber \\
 & \hspace{0.6cm} \xi_{1} \geq 0, \psi_{1} \geq 0, 
\end{align}}
and 
{
\begin{align}
\label{eq:10}
\underset{ w_{2}, b_{2}, \xi_{2},\psi_{2}}{min}  \hspace{0.5cm}~&\frac{1}{2}\|X_2 w_{2} + e_{2} b_{2}\|^2+c_{2} e_{1}^t \xi_{2} + c_{u}e_{u}^t \psi_{2} \nonumber \\
 \text { s.t. }\hspace{0.5cm}  & \left(X_1 w_{2} +e_{1}b_{2}\right) \geq e_{1}-\xi_{2}, \nonumber \\
 & -\left(U w_{2} +e_{u}b_{2}\right) \geq (-1+\epsilon)e_{u}-\psi_{2}, \nonumber \\
 & \hspace{0.6cm} \xi_{2} \geq 0, \psi_{2} \geq 0, 
\end{align}}
here, $c_i, (i=1,2,u)$ are positive parameters, $\xi_{1}, \xi_{2}, \psi_{1}, \psi_{2}$ are slacks variables.
The dual formulations of problems \eqref{eq:9} and \eqref{eq:10} can be obtaind as:
{
\begin{align}
\label{eq:11}
\underset{ \alpha_{1},\mu_{1}}{max}  \hspace{0.5cm}~&e_{2}^t\alpha_{1}-\frac{1}{2}(\alpha_{1}^tE-\mu_{1}^tF)(GG^t)^{-1}(E^t\alpha_{1}-F^t\mu_{1}) \nonumber \\
& +(\epsilon-1)e_{u}^t\mu_{1}  \nonumber \\
 \text { s.t }\hspace{0.5cm}  & 0\leq \mu_{1} \leq c_{u}, \nonumber \\
 & 0\leq \alpha_{1} \leq c_{1}, \nonumber \\ 
\end{align}}
and
{
\begin{align}
\label{eq:12}
\underset{ \alpha_{2},\mu_{2}}{max}  \hspace{0.5cm}~&e_{1}^t\alpha_{2}-\frac{1}{2}(\alpha_{2}^tG-\mu_{2}^tF)(EE^t)^{-1}(G^t\alpha_{2}-F^t\mu_{2}) \nonumber \\
& +(\epsilon-1)e_{u}^t\mu_{2}  \nonumber \\
 \text { s.t }\hspace{0.5cm}  & 0\leq \mu_{2} \leq c_{u}, \nonumber \\
 & 0\leq \alpha_{2} \leq c_{2}, \nonumber \\ 
\end{align}}
here, $G=[X_1, e_{1}]$, $E=[X_2,  e_{2}]$, $F=[U,  e_{u}]$ 
and $\alpha_{1}$, $\alpha_{2}$, $\mu_{1}$, $\mu_{2}$ are Lagrange multipliers of appropriate dimensions. 
The solution of \eqref{eq:11} and \eqref{eq:12} yields the separating hyperplane,
{
\begin{align}
& {\left[\begin{array}{l}
w_1 \\
b_1
\end{array}\right]=-\left(G^t G+\delta I\right)^{-1}\left(E^t \alpha_1-F^t \mu_1\right)} \\
& {\left[\begin{array}{l}
w_2 \\
b_2
\end{array}\right]=\left(E^t E+\delta I\right)^{-1}\left(G^t \alpha_2-F^t \mu_2\right)},
\end{align}}

Since the matrices $G^t G$ and $E^t E$ may be ill-conditioned, a small positive number, $\delta$, is added along the diagonal of $G^t G$ and $E^t E$ to ensure numerical stability $i.e. \hspace{0.2cm} G^t G+\delta I$  and $E^t E+\delta I$, where $I$ represents an identity matrix of suitable dimensions.

A new test instance $x$ can be classified by the following decision function:
\begin{align}
class \hspace{0.1cm} (i) = \underset{ i=1, 2}{\arg\min \hspace{0.1cm}}\frac{\|x w_{i} + b_{i}\|}{\|w_{i}\|}
\end{align}

\section{Proposed Intuitionistic Fuzzy Universum Twin Support Vector Machine for Imbalanced Data (IFUTSVM-ID)}
\label{Intuitionistic Fuzzy Universum Twin Support Vector Machine for Imbalanced Data}
In this section, we provide a detailed mathematical formulation of the proposed IFUTSVM-ID model tailored for linear and nonlinear cases. The proposed IF-edRVFL incorporates an intuitionistic fuzzy scheme, enabling the development of multiple diverse and robust base models. In case of class imbalance, the majority class holds a significant amount of information regarding the data distribution. When dealing with a majority-class hyperplane, we increase the number of data points in the universum data to proper information about the data. In contrast to UTSVM, which does not address the issue of class imbalance and considers the universum as not belonging to either of the two classes. Constraints on the universum samples are included such that their sizes are equivalent to the difference between the number of samples in the classes. To construct the hyperplane for the majority class, a constraint is applied to ensure that the distance of universum data points from the majority (negative) class must be $(1-\epsilon)$. In addition to giving prior information of data, the majority class hyperplane is not positioned too close to the data point of the minority (positive) class. Consequently, this scheme reduces the inclination of the classifier towards the majority class, leading to improved generalization performance.

Let $X_1 \in R^{ m_{1} \times n}$ and $X_2 \in R^{ m_{2} \times n}$ are two matrices of positive and negative classes, respectively. $X_2^*$ denotes the reduced data samples randomly selected from the negative class, with a size of $m_{1} \times n$. $U$ represents the universum data matrix of size $u \times n$, where $u=(m_{2}-m_{1})$ represents the count of data points that differ between the two classes. $U^*$ is a randomly chosen subset of $U$ consisting of $g= \lceil \frac{m_{1}}{2} \rceil$ data points. $e$ is vectors of ones of appropriate dimensions.

\subsection{Linear IFUTSVM-ID}

The optimization problem of the linear IFUTSVM-ID model is as follows:
{
\begin{align}
\label{eq:16}
\underset{ w_{1}, b_{1}, \xi_{1},\lambda_{1}}{min}  \hspace{0.5cm}~&\frac{1}{2}\|X_1 w_{1} + e_{1} b_{1}\|^2+\frac{c_{3}}{2}(\|w_{1}\|^2+b_{1}^2) \nonumber \\
& +c_{1} s_{2}^t \xi_{1} + c_{u}e_{g}^t \lambda_{1} \nonumber \\
 \text { s.t. }\hspace{0.5cm}  & -\left(X_2^* w_{1} +e_{2}b_{1}\right) \geq e_{2}-\xi_{1}, \nonumber \\
 & \left(U^* w_{1} +e_{g}b_{1}\right) \geq (-1+\epsilon)e_{g}-\lambda_{1}, \nonumber \\
 & \hspace{0.6cm} \xi_{1} \geq 0, \lambda_{1} \geq 0,
\end{align}}
and
{
\begin{align}
\label{eq:17}
\underset{ w_{2}, b_{2}, \xi_{2},\lambda_{2}}{min}  \hspace{0.5cm}~&\frac{1}{2}\|X_2 w_{2} + e_{2} b_{2}\|^2+\frac{c_{4}}{2}(\|w_{2}\|^2+b_{2}^2) \nonumber \\
&+c_{2} s_{1}^t \xi_{2} + c_{u}e_{d}^t \lambda_{2} \nonumber \\
 \text { s.t. }\hspace{0.5cm}  & \left(X_1 w_{2} +e_{1}b_{2}\right) \geq e_{1}-\xi_{2}, \nonumber \\
 & \left(U w_{2} +e_{d}b_{2}\right) \geq (\epsilon -1)e_{d}-\lambda_{2}, \nonumber \\
 & \hspace{0.6cm} \xi_{2} \geq 0, \lambda_{2} \geq 0, 
\end{align}}
here $c_{i}$ ($i=1, 2, 3, u$) denotes the tunable parameters, and the slack variables are denoted as $\xi_{1}$, $\lambda_{1}$, $\xi_{2}$, and $\lambda_{2}$. The score values corresponding to the negative and positive classes are represented by $s_{2} \in \mathbb{R}^{m_2}$ and $s_{1}\in \mathbb{R}^{m_1}$, respectively.

The Lagrangian of equation \eqref{eq:16} is written as:
{
\begin{align}
\label{eq:18}
 L(\Theta)& =\frac{1}{2}\|X_1 w_{1} + e_{1} b_{1}\|^2+\frac{c_{3}}{2}(\|w_{1}\|^2+b_{1}^2)+c_{1} s_{2}^t \xi_{1}   \nonumber \\
  & + c_{u}e_{g}^t \lambda_{1} - \alpha^t(-\left(X_2^* w_{1} +e_{2}b_{1}\right)-e_{2} + \xi_{1})\nonumber \\
 & \hspace{-1cm} - \beta^t (\left(U^* w_{1} +e_{g}b_{1}\right) - (-1+\epsilon)e_{g} + \lambda_{1}) - \mu^t\xi_{1} - \gamma^t\lambda_{1},
\end{align}}
where $\Theta = \{ w_{1},b_{1},\xi_{1},\lambda_{1},\alpha,\beta,\mu,\gamma\}$ and the Lagrange multipliers are denoted as $\alpha$, $\beta$, $\mu$ and $\gamma$.

By employing the K.K.T. conditions to equation \eqref{eq:18}, we obtain
\begin{align}
    X_1^t(X_1w_{1}+e_{1}b_{1})+ c_{3}w_{1} + X_2^{*t}\alpha - U^{*t}\beta = 0,\label{eqn:19} & \\
   e_{1}^t(X_1w_{1}+e_{1}b_{1})+ c_{3}b_{1} + e_{2}^t \alpha - e_{g}^t \beta = 0,\label{eqn:20} &\\
   c_{1}s_{2} - \alpha - \mu = 0, \label{eqn:21} & \\
   c_{u}e_{g} - \beta - \psi=0, \label{eqn:22}& \\
   \alpha^t(-\left(X_2^* w_{1} +e_{2}b_{1}\right)-e_{2} + \xi_{1})=0, &\\
   \beta^t (\left(U^* w_{1} +e_{d}b_{1}\right) - (-1+\epsilon)e_{g} + \lambda_{1}) = 0, & \\
   \mu^t\xi_{1} = 0, & \\
   \gamma^t\lambda_{1} = 0 \label{eq:AA}. 
\end{align}

Since $\mu >0$ and $\gamma > 0,$ \eqref{eqn:21} and \eqref{eqn:22} reduce to,
\begin{align*}
    0 \le \alpha \le c_{1}s_{2},\\
    0 \le \beta \le c_{u}e_{g}.
\end{align*}
Rewriting \eqref{eqn:19} and \eqref{eqn:20}, we obtain
\begin{align}
\label{eq:27}
    \binom{X_1^t}{e_{1}^t}\left(X_1 \hspace{0.5cm} e_{1}\right) \binom{w_{1}}{b_{1}} +c_{3}\binom{w_{1}}{b_{1}} + \binom{X_2^{*t}}{e_{2}^t}\alpha - \binom{U^{*t}}{e_{g}^t}=0.
\end{align}

Let $E=\left(X_1 \hspace{0.5cm} e_{1}  \right)$, $F^* = \left(X_2^{*} \hspace{0.5cm} e_{2}  \right)$ and $G^* = \left(U^{*} \hspace{0.5cm} e_{g}  \right)$, then rewrite \eqref{eq:27} as
\begin{align}
\label{eq:28}
    &\left(E^tE + c_{3}I\right)\binom{w_{1}}{b_{1}} + F^{*t} \alpha - G^{*t} \beta  \nonumber \\
  \text{i.e.}, \hspace{0.2cm}  &\binom{w_{1}}{b_{1}} = -\left(  E^tE+c_{3}I\right)^{-1} \left( F^{*t} \alpha - G^{*t} \beta \right)
\end{align}

With the help of equation \eqref{eq:28} and the K.K.T conditions \eqref{eqn:19} - \eqref{eq:AA}, the dual form of equation \eqref{eq:16} can be obtained as follows:
\begin{align}
\label{eq:29}
    \underset{\alpha, \beta}{max}  \hspace{0.5cm}~& -\frac{1}{2}\left( \alpha^tF^* - \beta^tG^* \right)(E^tE+c_{3}I)^{-1}(F^{*t}\alpha - G^{*t}\beta) \nonumber \\
    & + (\epsilon-1)e_{g}^t\beta + e_{1}^t\alpha \nonumber \\
     \text { s.t. }\hspace{0.5cm}  & 0 \le \alpha \le c_{1}s_{2}, \nonumber \\
     & 0 \le \beta \le c_{u}e_{g},
\end{align}
and
\begin{align}
\label{eq:30}
    \underset{\eta, \theta}{max}  \hspace{0.5cm}~& -\frac{1}{2}\left( \eta^tE + \theta^tG \right)(F^tF+c_{4}I)^{-1}(E^{t}\eta + G^{t}\theta)  \nonumber \\
    & + (1-\epsilon)e_{g}^t\theta + e_{1}^t\eta \nonumber \\
     \text { s.t. }\hspace{0.5cm}  & 0 \le \eta \le c_{2}s_{1}, \nonumber \\
     & 0 \le \theta \le c_{u}e_{d},
\end{align}
where, $F=\left(X_1 \hspace{0.5cm} e_{1}  \right)$, $G = \left(U \hspace{0.5cm} e_{d}  \right)$.

After solving \eqref{eq:29} and \eqref{eq:30}, we can acquire the separating hyperplanes as follows
\begin{align}
\binom{w_{1}}{b_{1}} = -\left(  E^tE+c_{3}I\right)^{-1} \left( F^{*t} \alpha - G^{*t} \beta \right)
\end{align}
\begin{align}
\label{eq:130}
    \binom{w_{2}}{b_{2}} = \left(  F^tF+c_{4}I\right)^{-1} \left( E^{t} \eta + G^{t} \theta \right)
\end{align}
The categorization of a new input data point $x$ into either the negative or positive class can be determined as follows:

{
\begin{align}
\label{eq:131}
class \hspace{0.05cm} (i) = \underset{ i=1, 2}{\arg\min \hspace{0.1cm}}\frac{\|x w_{i} + b_{i}\|}{\|w_{i}\|}
\end{align}}

\subsection{Non-linear IFUTSVM-ID}
In order to solve nonlinear classification problems, one might contemplate using the following kernel function:
\begin{align*}
    \mathscr{K}(x,D^T)w_1+b_1=0,   \hspace{0.2cm} \mathscr{K}(x,D^T)w_2+b_2=0 
\end{align*}
where $\mathscr{K}(x_1, x_2) = (\psi(x_1), \psi(x_2))$ is a kernel function and $D = [X_1; X_2]$. The optimization problem of non-linear IFUTSVM-ID model is as follows:
\begin{align}
\label{eq:34}
\underset{ w_{1}, b_{1}, \xi_{1},\lambda_{1}}{min}  \hspace{0.5cm}~&\frac{1}{2}\|\mathscr{K}(X_1,D^t) w_{1} + e_{1} b_{1}\|^2+\frac{c_{3}}{2}(\|w_{1}^2\|+b_{1}^2) \nonumber \\
& +c_{1} s_{2}^t \xi_{1} + c_{u}e_{g}^t \lambda_{1} \nonumber \\
 \text { s.t. }\hspace{0.5cm}  & -\left(\mathscr{K}(X_2^*,D^t) w_{1} +e_{2}b_{1}\right) \geq e_{2}-\xi_{1}, \nonumber \\
 & \left(\mathscr{K}(U^*,D^t) w_{1} +e_{g}b_{1}\right) \geq (-1+\epsilon)e_{g}-\lambda_{1}, \nonumber \\
 & \hspace{0.6cm} \xi_{1} \geq 0, \lambda_{1} \geq 0,
\end{align}
and
\begin{align}
\label{eq:35}
\underset{ w_{2}, b_{2}, \xi_{2},\lambda_{2}}{min}  \hspace{0.5cm}~&\frac{1}{2}\|\mathscr{K}(X_2,D^t) w_{2} + e_{2} b_{2}\|^2+\frac{c_{4}}{2}(\|w_{2}^2\|+b_{2}^2) \nonumber \\
& +c_{2} s_{1}^t \xi_{2} + c_{u}e_{d}^t \lambda_{2} \nonumber \\
 \text { s.t. }\hspace{0.5cm}  & \left(\mathscr{K}(X_1,D^t) w_{2} +e_{1}b_{2}\right) \geq e_{1}-\xi_{2}, \nonumber \\
 & \left(\mathscr{K}(U,D^t) w_{2} +e_{d}b_{2}\right) \geq (\epsilon -1)e_{d}-\lambda_{2}, \nonumber \\
 & \hspace{0.6cm} \xi_{2} \geq 0, \lambda_{2} \geq 0, 
\end{align}
here, $c_{i}$ ($i=1,2,3,4,u$) denotes the parameters, and the slack variables are denoted as $\xi_{1}$, $\xi_{2}$, $\lambda_{1}$, and $\lambda_{2}$. The score values for the positive and negative classes are given by $s_{1} \in \mathbb{R}^{m_1}$ and $s_{2} \in \mathbb{R}^{m_2}$, respectively.

The Wolfe duals of the problems of \eqref{eq:34} and \eqref{eq:35} can be obtained as:
\begin{align}
\label{eq:36}
    \underset{\alpha, \beta}{max}  \hspace{0.5cm}~& -\frac{1}{2}\left( \alpha^tF^* - \beta^tG^* \right)(E^tE+c_{3}I)^{-1}(F^{*t}\alpha - G^{*t}\beta) \nonumber \\
    & + (\epsilon-1)e_{g}^t\beta + e_{1}^t\alpha \nonumber \\
     \text { s.t. }\hspace{0.5cm}  & 0 \le \alpha \le c_{1}s_{2}, \nonumber \\
     & 0 \le \beta \le c_{u}e_{g},
\end{align}
and
\begin{align}
\label{eq:37}
    \underset{\eta, \theta}{max}  \hspace{0.5cm}~& -\frac{1}{2}\left( \eta^tE + \theta^tG \right)(F^tF+c_{4}I)^{-1}(E^{t}\eta + G^{t}\theta)  \nonumber \\
    & + (1-\epsilon)e_{g}^t\theta + e_{1}^t\eta \nonumber \\
     \text { s.t. }\hspace{0.5cm}  & 0 \le \eta \le c_{2}s_{1}, \nonumber \\
     & 0 \le \theta \le c_{u}e_{d},
\end{align}
where, $E=\left(\mathscr{K}(X_1, D^t), \hspace{0.2cm} e_{1}  \right)$, $F=\left(\mathscr{K}(X_1, D^t), \hspace{0.2cm} e_{1}  \right)$, $G = \left(\mathscr{K}(U, D^t), \hspace{0.2cm} e_{d}  \right)$, $F^* = \left(\mathscr{K}(X_2^{*},D^t), \hspace{0.2cm} e_{2}  \right)$, $G^* = \left(\mathscr{K}(U^{*}, D^t), \hspace{0.2cm} e_{g}  \right)$.

After solving \eqref{eq:36} and \eqref{eq:37}, the separating hyperplanes can be attained as,
\begin{align}
\label{eq:136}
\binom{w_{1}}{b_{1}} = -\left(  E^tE + c_{3}I\right)^{-1} \left( F^{*t} \alpha - G^{*t} \beta \right)
\end{align}
\begin{align}
\label{eq:137}
    \binom{w_{2}}{b_{2}} = \left(  F^tF + c_{4}I\right)^{-1} \left( E^{t} \eta + G^{t} \theta \right)
\end{align}
The following decision function can be utilized to predict the label of a new sample $x$.

\begin{align}
\label{eq:138}
class \hspace{0.05cm} (i) = \underset{ i=1, 2} {\arg\min} \hspace{0.1cm} \frac{\| \mathscr{K}(x^t,D^t)w_{i} + b_{i}\|}{\|w_{i}\|}.
\end{align}
The algorithm and time complexity analysis of the proposed IFUTSVM-ID model are detailed in Section S.II of the supplementary material.

\section{Experimental results}
\label{Experimental results}
To test the efficiency of proposed IFUTSVM-ID  model, we compare their performance against baseline models, including UTSVM \cite{qi2012twin}, RUTSVM-CIL \cite{richhariya2020reduced}, IF-USVM \cite{kumari2023intuitionistic}, IFW-LSTSVM \cite{tanveer2022intuitionistic}, and CGFTSVM-ID \cite{anuradha2024}. We utilize the publicly available KEEL \cite{derrac2015keel} benchmark datasets under different noise levels. Furthermore, to assess the effectiveness of the proposed IFUTSVM-ID model, we utilized it for the diagnosis of Alzheimer's Disease (AD). AD dataset is accessible through the Alzheimer’s Disease Neuroimaging Initiative (ADNI) (adni.loni.usc.edu). In the Supplementary Material, we discuss the experimental setup and the experimental results on KEEL datasets with label noise.

\begin{table*}[]
\centering
    \caption{Average accuracy (ACC) and average rank of the proposed IFUTSVM-ID model with the baseline models based on KEEL datasets.}
    \label{Classification performance on KEEL dataset.}
    \resizebox{1\textwidth}{!}{
\begin{tabular}{lcccccc}
\hline
Metric $\downarrow$ \vline~ Model $\rightarrow$ & UTSVM \cite{qi2012twin}& RUTSVM-CIL \cite{richhariya2020reduced}&  IF-USVM \cite{kumari2023intuitionistic}& IFW-LSTSVM \cite{tanveer2022intuitionistic}& CGFTSVM-ID \cite{anuradha2024}& IFUTSVM-ID \\ 
 \hline
Average ACC & $83.52$ & $81.15$ & $80.61$ & $79.44$ & $86.19$ & $87.70$ \\ \hline
Average Rank & $3.71$ & $4.24$ & $4.47$ & $4.33$ & $2.76$ & $1.54$ \\ \hline
\end{tabular}}
\end{table*}

\subsection{Evaluation on KEEL datasets}
In this subsection, we present an intricate analysis involving a comparison of the proposed IFUTSVM-ID model against baseline UTSVM, RUTSVM-CIL, IF-USVM, IFW-LSTSVM, and CGFTSVM-ID models. This comparison is carried out across $46$ KEEL benchmark datasets spanning diverse domains. Table S.II presents the classification ACC, along with the optimal parameters of the proposed IFUTSVM-ID model compared to the existing models in the Supplementary material. Table \ref{Classification performance on KEEL dataset.} shows the average ACC and the average Rank of the proposed IFUTSVM-ID model compared to the existing models. The average ACC of the existing UTSVM, RUTSVM-CIL, IF-USVM, IFW-LSTSVM, and CGFTSVM-ID models are $83.52\%$, $81.15\%$, $80.61\%$, $79.44\%$, and $86.19\%$ respectively, whereas the average ACC of the proposed IFUTSVM-ID model is $87.70\%$. The proposed IFUTSVM-ID secured the top position in terms of average ACC. This observation strongly emphasizes the significant superiority of the proposed models over baseline models. Depending solely on average ACC as a single metric could pose problems, as outstanding performance on specific datasets may obscure insufficient performance on others. To address this concern, it becomes crucial to rank each model individually with respect to each dataset, enabling a thorough evaluation of their respective capabilities. In the ranking scheme \cite{demvsar2006statistical}, the model with the lowest performance on a dataset receives a higher rank, whereas the model achieving the best performance is assigned a lower rank. In the evaluation of $p$ models across $N$ datasets, the rank of the $j^{th}$ model on the $i^{th}$ dataset can be denoted as $R_j^i$. Then, $R_j = \frac{1}{N}\sum_{i=1}^NR_j^i$ determined the average rank of the model. The average rank of the proposed IFUTSVM-ID with the existing UTSVM, RUTSVM-CIL, IF-USVM, IFW-LSTSVM, and CGFTSVM-ID models are $3.71$, $4.24$, $4.47$, $4.33$, $2.76$ and $1.54$, respectively. The proposed IFUTSVM-ID model attained the lowest average rank among all the models. Since a lower rank indicates a better-performing model, the proposed IFUTSVM-ID emerged as the top-performing model. We now proceed to conduct statistical tests to determine the significance of the results. Firstly, we employ the Friedman test \cite{demvsar2006statistical} to determine whether there are significant differences among the models. Under the null hypothesis, it is assumed that the average rank of the models is equivalent, indicating equal performance. The Friedman statistic is given by: $\chi_{F}^2 = \frac{12N}{p(p+1)}\left[ \sum_{i=1} R_{j}^2 - \frac{p(p+1)^2}{4}\right]$. However, the Friedman statistic tends to be overly cautious. To address this, the better statistic \cite{iman1980approximations} is given as follows: $F_{F}= \frac{(N-1)\chi_{F}^2}{N(p-1)-\chi_{F}^2}$, where $F_F$ follows an $F$-distribution with d.o.f $((p-1), (p-1)(N-1))$. For $p=6$ and $N=46$, we get $\chi_F^2 = 91.48$ and $F_F = 29.72$. From the $F$-distribution table $F_F (5, 225) = 2.2542$ at a $5\%$ level of significance. The null hypothesis rejected as $F_F > 2.2542$. Therefore, significant disparities exist among the models. Subsequently, we utilize the Nemenyi post hoc test \cite{demvsar2006statistical} to scrutinize the pairwise distinctions between the models. The critical difference $(C.D.)$ is computed at a significance level of $5\%$ as $C.D. = q_\alpha \sqrt{\frac{p(p+1)}{6N}}$, where $q_{\alpha}$ denotes the critical value obtained from the distribution table for the two-tailed Nemenyi test. Referring to the statistical $F$-distribution table, at a significance level of $5\%$, we find $q_{\alpha} = 2.850$, which leads to a critical difference of $C.D. = 1.11$. The average differences in rank between the proposed IFUTSVM-ID model and the baseline UTSVM, RUTSVM-CIL, IF-USVM, IFW-LSTSVM, and CGFTSVM-ID models are $2.17$, $2.70$, $2.93$, $2.79$, and $2.76$, respectively. The Nemenyi post hoc test reveals significant differences between the proposed IFUTSVM-ID model and the baseline models. Considering all these findings, we infer that the proposed IFUTSVM-ID model demonstrates superior performance when compared to the baseline models.

\begin{figure*}[ht!]
\begin{minipage}{.246\linewidth}
\centering
\subfloat[abalone9-18]{\label{fig3:1a}\includegraphics[scale=0.25]{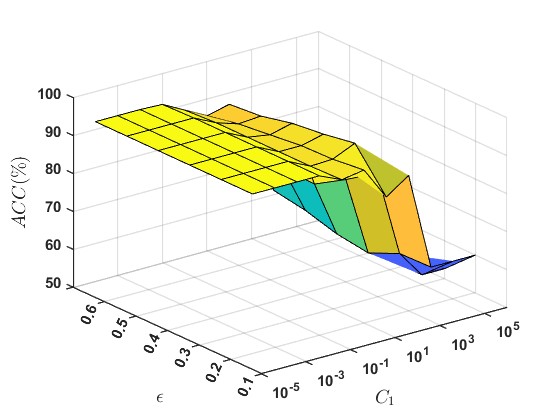}}
\end{minipage}
\begin{minipage}{.246\linewidth}
\centering
\subfloat[aus]{\label{fig3:1b}\includegraphics[scale=0.25]{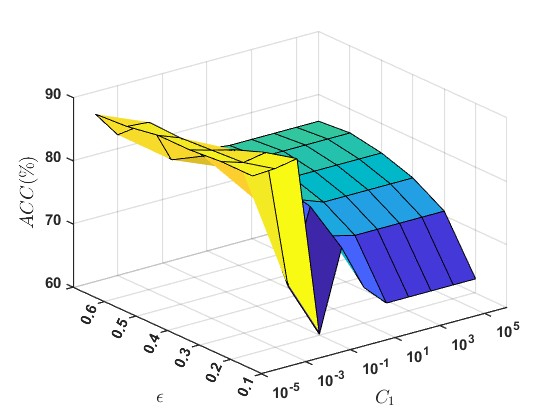}}
\end{minipage}
\begin{minipage}{.246\linewidth}
\centering
\subfloat[ecoli-0-1-4-6\_vs\_5]{\label{fig3:11d}\includegraphics[scale=0.25]{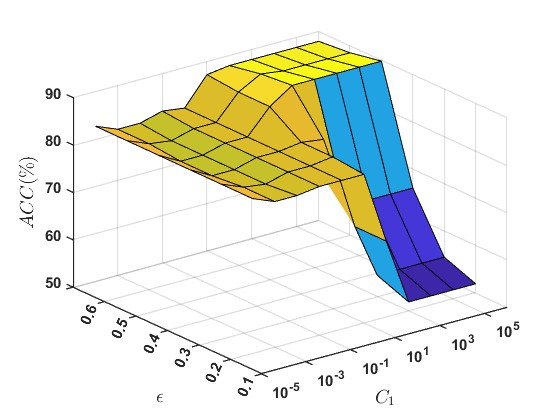}}
\end{minipage}
\begin{minipage}{.246\linewidth}
\centering
\subfloat[votes]{\label{fig3:1e}\includegraphics[scale=0.25]{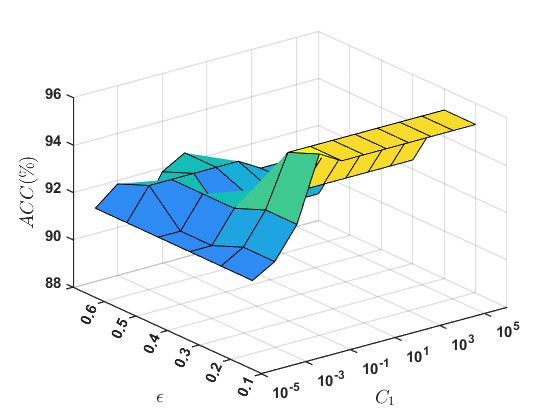}}
\end{minipage}
\caption{Effect of parameters $C_1$ and $\epsilon$ on the performance of the proposed IFUTSVM-ID  model.}
\label{Effect of parameters C and eps}
\end{figure*}

Based on the above analysis, it is evident that the proposed IFUTSVM-ID model performs better than the existing models. The hyperparameters of the proposed IFUTSVM-ID model significantly affect its generalization performance. Figure \ref{Effect of parameters C and eps} displays the sensitivity of the parameters of the proposed IFUTSVM-ID model. Therefore, it is crucial to meticulously select the hyperparameters for the proposed IFUTSVM-ID models to achieve the best possible generalization performance. 

\subsection{Evaluation on ADNI dataset}
Alzheimer's disease (AD) is a progressive neurological disorder that detrimentally affects memory and cognitive functions \cite{tanveer2024ensemble}. AD typically initiates with mild cognitive impairment (MCI). In order to train the proposed IFUTSVM-ID model, ADNI repository ($adni.loni.usc.edu$) scans dataset is used. The ADNI project, spearheaded by Michael W. Weiner in $2003$, is designed to investigate various neuroimaging techniques, such as positron emission tomography (PET), magnetic resonance imaging (MRI), and other diagnostic tests for AD, particularly focusing on the MCI stage. The feature extraction pipeline chosen in this paper follows the same approach as outlined in \cite{richhariya2021efficient}. The dataset comprises three scenarios: MCI versus AD, control normal (CN) versus MCI, and CN versus AD.

\begin{table*}[htp]
\centering
\caption{Performance comparison of the proposed IFUTSVM-ID along with the baseline models based on classification ACC on AD and BC datasets.}
    \label{Classification accuracy biomedical}
\resizebox{1.0\textwidth}{!}{
\begin{tabular}{lcccccc}
\hline
Dataset & UTSVM \cite{qi2012twin} & RUTSVM-CIL \cite{richhariya2020reduced} &  IF-USVM  \cite{kumari2023intuitionistic} & IFW-LSTSVM \cite{tanveer2022intuitionistic} & CGFTSVM-ID \cite{anuradha2024} &  IFUTSVM-ID \\
 & $(ACC (\%), Sensitivity)$ & $(ACC (\%), Sensitivity)$ & $(ACC (\%), Sensitivity)$ & $(ACC (\%), Sensitivity)$ & $(ACC (\%), Sensitivity)$ & $(ACC (\%), Sensitivity)$ \\
 & $(Specificity, Precision)$ & $(Specificity, Precision)$ & $(Specificity, Precision)$ & $(Specificity, Precision)$ & $(Specificity, Precision)$ & $(Specificity, Precision)$ \\ \hline
CN\_vs\_AD & $(89.52, 81.48)$ & $(79.03, 92.59)$ & $(83.87, 72.22)$ & $(49.19, 55.94)$ & $(89.13, 89.32)$ & $(89.87, 72.22)$ \\
 & $(90.71, 75.62)$ & $(68.57, 69.44)$ & $(92.86, 88.64)$ & $(47.14, 44.94)$ & $(90.3, 86.34)$ & $(91.43, 86.67)$ \\
CN\_vs\_MCI & $(66.85, 79.83)$ & $(68.59, 68.56)$ & $(58.29, 47.37)$ & $(50.8, 57.01)$ & $(67.59, 76.39)$ & $(70.85, 74.04)$ \\
 & $(46.58, 70)$ & $(65.36, 64.89)$ & $(75.34, 75)$ & $(49.92, 61)$ & $(55.81, 76.39)$ & $(55.84, 75.04)$ \\
MCI\_vs\_AD & $(72, 37.5)$ & $(53.14, 58.93)$ & $(66.29, 58.93)$ & $(68, 53.51)$ & $(73.71, 46.51)$ & $(72, 51.79)$ \\
 & $(58.24, 60)$ & $(50.42, 35.87)$ & $(57.75, 47.83)$ & $(53.26, 50.89)$ & $(57.2, 46.51)$ & $(58.82, 47.18)$ \\ \hline
Average ACC & $76.12$ & $66.92$ & $69.48$ & $56$ & $76.81$ & $77.57$ \\ \hline
\end{tabular}}
\end{table*}

\begin{figure*}[htp]
\begin{minipage}{.328\linewidth}
\centering
\subfloat[CN\_vs\_AD]{\label{fig2:1a}\includegraphics[scale=0.27]{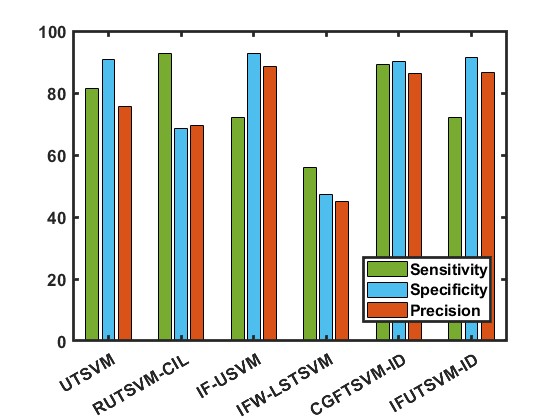}}
\end{minipage}
\begin{minipage}{.328\linewidth}
\centering
\subfloat[CN\_vs\_MCI]{\label{fig2:1b}\includegraphics[scale=0.27]{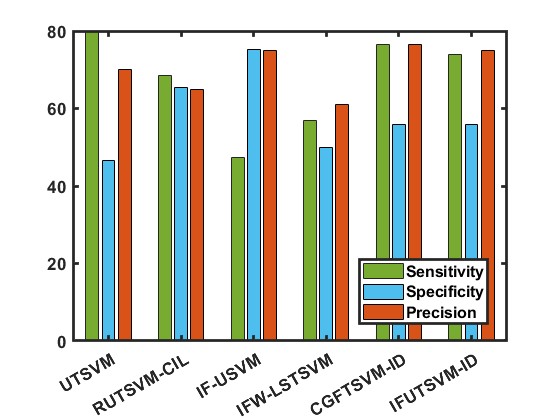}}
\end{minipage}
\begin{minipage}{.328\linewidth}
\centering
\subfloat[MCI\_vs\_AD]{\label{fig2:1c}\includegraphics[scale=0.27]{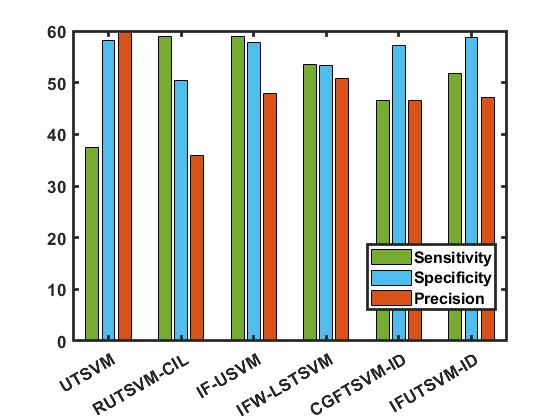}}
\end{minipage}
\caption{Performance comparison of the proposed IFUTSVM-ID model along with the baseline models on ADNI data using sensitivity, specificity, and precision.}
\label{Effect of parameters}
\end{figure*}

The ACC values for the proposed IFUTSVM-ID model along with the existing models for AD diagnosis, are presented in Table \ref{Classification accuracy biomedical}. With an average ACC of $77.57\%$, the proposed IFUTSVM-ID emerges as the top classifier in our analysis. The average ACC of the baseline UTSVM, RUTSVM-CIL, IF-USVM, IFW-LSTSVM, and CGFTSVM-ID models are $76.12$, $66.92$, $69.48$, $56$, and $76.81$, respectively. Compared to the second-top model, CGFTSVM-ID, the proposed IFUTSVM-ID model exhibits an average ACC that is approximately $0.76\%$ higher. The proposed IFUTSVM-ID achieves the highest ACC of $89.87\%$ for the CN vs AD case. In the CN vs MCI case, the proposed IFUTSVM-ID emerges as the top performer, with average accuracies of $70.85\%$. Therefore, the proposed IFUTSVM-ID consistently demonstrates superior performance by achieving high ACC across various cases, establishing its prominence among the models. Furthermore, to illustrate the comparison between the proposed IFUTSVM-ID model and the baseline UTSVM, RUTSVM-CIL, IF-USVM, IFW-LSTSVM, and CGFTSVM-ID models in terms of specificity, sensitivity, and precision, we plotted the bar graphs shown in Figure \ref{Effect of parameters}. The performance of the proposed IFUTSVM-ID model is almost comparable to that of the baseline models. The overall findings emphasize the effectiveness of the proposed models in discerning various cognitive states.

\section{Conclusion}
\label{Conclusion}
In this paper, we proposed the intuitionistic fuzzy universum twin support vector machine for imbalanced data (IFUTSVM-ID) to enhance their robustness against noise and outliers present in the dataset. Unlike UTSVM, which employs a uniform weighting approach to generate the optimal classifier, the proposed IFUTSVM-ID model employs an intuitionistic fuzzy membership scheme to compute the final parameters. As a result, the proposed IFUTSVM-ID achieves superior generalization performance by incorporating prior information into the training process. The approach integrating undersampling with oversampling using universum data has been found to be beneficial in classifying imbalanced datasets. One of the key advantages of incorporating a regularization term into the primal optimization formulation of IFUTSVM-ID is its adherence to the SRM principle. This approach bypasses the need for matrix inversions, thereby streamlining computational processes. Furthermore, it effectively mitigates overfitting concerns, enhancing the model's generalization capabilities.
We evaluate the proposed IFUTSVM-ID model on $46$ KEEL datasets to experimentally validate the performance. Statistical evaluations, including metrics such as accuracy (ACC), ranking scheme analysis, Win-tie-loss sign test, Friedman test, and Nemenyi post hoc test, conclusively demonstrate that the proposed IFUTSVM-ID model exhibits superior performance compared to other classifiers. Also, we employed the proposed IFUTSVM-ID model for the diagnosis of Alzheimer’s disease datasets. Upon analysis, it has been ascertained that the proposed IFUTSVM-ID model demonstrates the highest level of classification performance. In future, there is a possibility of examining the selection of universum data as it has an impact on the performance of the model. Further, to enhance the performance of our IFUTSVM-ID model, various undersampling and oversampling techniques can be employed.

\section*{Acknowledgment}
This study receives support from the Science and Engineering Research Board (SERB) through the Mathematical Research Impact-Centric Support (MATRICS) scheme Grant No. MTR/2021/000787. 
The dataset used in this study was made possible through the Alzheimer’s Disease Neuroimaging Initiative (ADNI), funded by the Department of Defense (contract W81XWH-12-2-0012) and the National Institutes of Health (grant U01 AG024904). Support for ADNI came from the National Institute on Aging, the National Institute of Biomedical Imaging and Bioengineering, and numerous contributions from organizations:  Bristol-Myers Squibb Company; Transition Therapeutics Elan Pharmaceuticals, Inc.; IXICO Ltd.; Lundbeck;  Johnson \& Johnson Pharmaceutical Research \& Development LLC.;  Araclon Biotech; Janssen Alzheimer Immunotherapy Research \& Development, LLC.;  Novartis Pharmaceuticals Corporation; Eisai Inc.; Neurotrack Technologies; AbbVie, Alzheimer’s Association; CereSpir, Inc.; Lumosity; Biogen; Fujirebio; EuroImmun; Piramal Imaging; GE Healthcare; Cogstate; Meso Scale Diagnostics, LLC.; NeuroRx Research; Alzheimer’s Drug Discovery Foundation;  Servier; Eli Lilly and Company;  BioClinica, Inc.; Merck \& Co., Inc.; F. Hoffmann-La Roche Ltd. and its affiliated company Genentech, Inc.;   Pfizer Inc. and Takeda Pharmaceutical Company. The Canadian Institutes of Health Research supported ADNI’s clinical sites across Canada. Private sector donations were facilitated through the Foundation for the National Institutes of Health (www.fnih.org). The Northern California Institute and the Alzheimer's Therapeutic Research Institute at the University of Southern California provided support for research and teaching awards. The Neuro Imaging Laboratory at the University of Southern California facilitated the public release of the ADNI initiative's data. This investigation utilizes the ADNI dataset, which is accessible at \url{adni.loni.usc.edu}.

\bibliographystyle{IEEEtranN}
\bibliography{refs.bib}

\begin{thebibliography}{40}
\providecommand{\natexlab}[1]{#1}
\providecommand{\url}[1]{#1}
\csname url@samestyle\endcsname
\providecommand{\newblock}{\relax}
\providecommand{\bibinfo}[2]{#2}
\providecommand{\BIBentrySTDinterwordspacing}{\spaceskip=0pt\relax}
\providecommand{\BIBentryALTinterwordstretchfactor}{4}
\providecommand{\BIBentryALTinterwordspacing}{\spaceskip=\fontdimen2\font plus
\BIBentryALTinterwordstretchfactor\fontdimen3\font minus
  \fontdimen4\font\relax}
\providecommand{\BIBforeignlanguage}[2]{{%
\expandafter\ifx\csname l@#1\endcsname\relax
\typeout{** WARNING: IEEEtranN.bst: No hyphenation pattern has been}%
\typeout{** loaded for the language `#1'. Using the pattern for}%
\typeout{** the default language instead.}%
\else
\language=\csname l@#1\endcsname
\fi
#2}}
\providecommand{\BIBdecl}{\relax}
\BIBdecl

\bibitem[Cortes and Vapnik(1995)]{cortes1995support}
C.~Cortes and V.~Vapnik, ``Support-vector networks,'' \emph{Machine Learning},
  vol.~20, pp. 273--297, 1995.

\bibitem[Zhang et~al.(2017, doi: https://doi.org/10.1155/2017/8092691)Zhang,
  Jiang, Han, Wang, Yang, and Yang]{zhang2017rotating}
X.~Zhang, D.~Jiang, T.~Han, N.~Wang, W.~Yang, and Y.~Yang, ``Rotating machinery
  fault diagnosis for imbalanced data based on fast clustering algorithm and
  support vector machine,'' \emph{Journal of Sensors}, vol. 2017, 2017, doi:
  https://doi.org/10.1155/2017/8092691.

\bibitem[Joachims(2005)]{joachims2005text}
T.~Joachims, ``Text categorization with support vector machines: Learning with
  many relevant features,'' in \emph{Machine Learning: ECML-98: 10th European
  Conference on Machine Learning Chemnitz, Germany, April 21--23, 1998
  Proceedings}.\hskip 1em plus 0.5em minus 0.4em\relax Springer, 2005, pp.
  137--142.

\bibitem[Tomar and Agarwal(2015)]{tomar2015hybrid}
D.~Tomar and S.~Agarwal, ``Hybrid feature selection based weighted least
  squares twin support vector machine approach for diagnosing breast cancer,
  hepatitis, and diabetes,'' \emph{Advances in Artificial Neural Systems}, vol.
  2015, 2015.

\bibitem[Richhariya and Tanveer(2018{\natexlab{a}})]{richhariya2018eeg}
B.~Richhariya and M.~Tanveer, ``E{EG} signal classification using universum
  support vector machine,'' \emph{Expert Systems with Applications}, vol. 106,
  pp. 169--182, 2018.

\bibitem[Jayadeva et~al.(2007)Jayadeva, Khemchandani, and
  Chandra]{khemchandani2007twin}
Jayadeva, R.~Khemchandani, and S.~Chandra, ``Twin support vector machines for
  pattern classification,'' \emph{IEEE Transactions on Pattern Analysis and
  Machine Intelligence}, vol.~29, no.~5, pp. 905--910, 2007.

\bibitem[Shao et~al.(2011)Shao, Zhang, Wang, and Deng]{shao2011improvements}
Y.-H. Shao, C.-H. Zhang, X.-B. Wang, and N.-Y. Deng, ``Improvements on twin
  support vector machines,'' \emph{IEEE Transactions on Neural Networks},
  vol.~22, no.~6, pp. 962--968, 2011.

\bibitem[Quadir et~al.(2024, 10.1109/TCSS.2024.3411395)Quadir, Sajid, and
  Tanveer]{quadir2024granularGB}
A.~Quadir, M.~Sajid, and M.~Tanveer, ``Granular ball twin support vector
  machine,'' \emph{IEEE Transactions on Neural Networks and Learning Systems},
  2024, 10.1109/TCSS.2024.3411395.

\bibitem[Quadir and Tanveer(2024,
  10.1109/TCSS.2024.3411395)]{quadir2024granular}
A.~Quadir and M.~Tanveer, ``Granular ball twin support vector machine with
  pinball loss function,'' \emph{IEEE Transactions on Computational Social
  Systems}, 2024, 10.1109/TCSS.2024.3411395.

\bibitem[Tanveer et~al.(2016)Tanveer, Khan, and Ho]{tanveer2016robust}
M.~Tanveer, M.~A. Khan, and S.~S. Ho, ``Robust energy-based least squares twin
  support vector machines,'' \emph{Applied Intelligence}, vol.~45, pp.
  174--186, 2016.

\bibitem[Quadir and Tanveer(2024)]{quadir2024multiview}
A.~Quadir and M.~Tanveer, ``Multiview learning with twin parametric margin
  {SVM},'' \emph{Neural Networks}, vol. 180, p. 106598, 2024.

\bibitem[Quadir et~al.(2025,
  https://doi.org/10.1016/j.neunet.2025.107433)Quadir, Akhtar, and
  Tanveer]{quadir2024enhancing}
A.~Quadir, M.~Akhtar, and M.~Tanveer, ``Enhancing multiview synergy: {R}obust
  learning by exploiting the wave loss function with consensus and
  complementarity principles,'' \emph{Neural Networks}, p. 107433, 2025,
  https://doi.org/10.1016/j.neunet.2025.107433.

\bibitem[Quadir and Tanveer(2025)]{quadir2025trkm}
A.~Quadir and M.~Tanveer, ``{TRKM}: {T}win restricted kernel machines for
  classification and regression,'' \emph{arXiv preprint arXiv:2502.15759},
  2025.

\bibitem[Quadir et~al.(2025)Quadir, Sajid, and Tanveer]{quadir2025one}
A.~Quadir, M.~Sajid, and M.~Tanveer, ``One class restricted kernel machines,''
  \emph{arXiv preprint arXiv:2502.10443}, 2025.

\bibitem[Weston et~al.(2006)Weston, Collobert, Sinz, Bottou, and
  Vapnik]{weston2006inference}
J.~Weston, R.~Collobert, F.~Sinz, L.~Bottou, and V.~Vapnik, ``Inference with
  the universum,'' in \emph{Proceedings of the 23rd International Conference on
  Machine Learning}, 2006, pp. 1009--1016.

\bibitem[Qi et~al.(2012)Qi, Tian, and Shi]{qi2012twin}
Z.~Qi, Y.~Tian, and Y.~Shi, ``Twin support vector machine with universum
  data,'' \emph{Neural Networks}, vol.~36, pp. 112--119, 2012.

\bibitem[Zhao et~al.(2019)Zhao, Xu, and Fujita]{zhao2019improved}
J.~Zhao, Y.~Xu, and H.~Fujita, ``An improved non-parallel universum support
  vector machine and its safe sample screening rule,'' \emph{Knowledge-Based
  Systems}, vol. 170, pp. 79--88, 2019.

\bibitem[Moosaei and Hlad{\'\i}k(2023)]{moosaei2023lagrangian}
H.~Moosaei and M.~Hlad{\'\i}k, ``A lagrangian-based approach for universum twin
  bounded support vector machine with its applications,'' \emph{Annals of
  Mathematics and Artificial Intelligence}, vol.~91, no.~2, pp. 109--131, 2023.

\bibitem[Lou and Xie(2024, doi:
  https://doi.org/10.1016/j.eswa.2024.123480)]{lou2024multi}
C.~Lou and X.~Xie, ``Multi-view universum support vector machines with
  insensitive pinball loss,'' \emph{Expert Systems with Applications}, p.
  123480, 2024, doi: https://doi.org/10.1016/j.eswa.2024.123480.

\bibitem[Hazarika et~al.(2023)Hazarika, Gupta, and Kumar]{hazarika2023eeg}
B.~B. Hazarika, D.~Gupta, and B.~Kumar, ``E{EG} signal classification using a
  novel universum-based twin parametric-margin support vector machine,''
  \emph{Cognitive Computation}, pp. 1--16, 2023.

\bibitem[Quadir et~al.(2024)Quadir, Ganaie, and
  Tanveer]{quadir2024intuitionistic}
A.~Quadir, M.~A. Ganaie, and M.~Tanveer, ``Intuitionistic fuzzy generalized
  eigenvalue proximal support vector machine,'' \emph{Neurocomputing}, vol.
  608, p. 128258, 2024.

\bibitem[Akhtar et~al.(2024)Akhtar, Quadir, Tanveer, and
  Arshad]{akhtar2024flexi}
M.~Akhtar, A.~Quadir, M.~Tanveer, and M.~Arshad, ``Flexi-fuzz least squares
  {SVM} for {A}lzheimer's diagnosis: {T}ackling noise, outliers, and class
  imbalance,'' \emph{arXiv preprint arXiv:2410.14207}, 2024.

\bibitem[Fan et~al.(2017)Fan, Wang, Li, Gao, and Zha]{fan2017entropy}
Q.~Fan, Z.~Wang, D.~Li, D.~Gao, and H.~Zha, ``Entropy-based fuzzy support
  vector machine for imbalanced datasets,'' \emph{Knowledge-Based Systems},
  vol. 115, pp. 87--99, 2017.

\bibitem[Richhariya and Tanveer(2019)]{richhariya2019fuzzy}
B.~Richhariya and M.~Tanveer, ``A fuzzy universum support vector machine based
  on information entropy,'' in \emph{Machine Intelligence and Signal
  Analysis}.\hskip 1em plus 0.5em minus 0.4em\relax Springer, 2019, pp.
  569--582.

\bibitem[Li and Ma(2013)]{li2013fuzzy}
K.~Li and H.~Ma, ``A fuzzy twin support vector machine algorithm,''
  \emph{International Journal of Application or Innovation in Engineering and
  Management}, vol.~2, no.~3, pp. 459--465, 2013.

\bibitem[Richhariya and Tanveer(2018{\natexlab{b}})]{richhariya2018robust}
B.~Richhariya and M.~Tanveer, ``A robust fuzzy least squares twin support
  vector machine for class imbalance learning,'' \emph{Applied Soft Computing},
  vol.~71, pp. 418--432, 2018.

\bibitem[Xu et~al.(2019)Xu, Zhang, Zhao, Yang, and Pan]{xu2019knn}
Y.~Xu, Y.~Zhang, J.~Zhao, Z.~Yang, and X.~Pan, ``K{NN}-based maximum margin and
  minimum volume hyper-sphere machine for imbalanced data classification,''
  \emph{International Journal of Machine Learning and Cybernetics}, vol.~10,
  pp. 357--368, 2019.

\bibitem[Atanassov and Atanassov(1999, doi:
  https://doi.org/10.1007/978-3-7908-1870-3\_1)]{atanassov1999intuitionistic}
K.~T. Atanassov and K.~T. Atanassov, \emph{Intuitionistic fuzzy sets}.\hskip
  1em plus 0.5em minus 0.4em\relax Springer, 1999, doi:
  https://doi.org/10.1007/978-3-7908-1870-3\_1.

\bibitem[Rezvani et~al.(2019)Rezvani, Wang, and
  Pourpanah]{rezvani2019intuitionistic}
S.~Rezvani, X.~Wang, and F.~Pourpanah, ``Intuitionistic fuzzy twin support
  vector machines,'' \emph{IEEE Transactions on Fuzzy Systems}, vol.~27,
  no.~11, pp. 2140--2151, 2019.

\bibitem[Richhariya et~al.(2021)Richhariya, Tanveer, and {Alzheimer’s Disease
  Neuroimaging Initiative}]{richhariya2021efficient}
B.~Richhariya, M.~Tanveer, and {Alzheimer’s Disease Neuroimaging Initiative},
  ``An efficient angle-based universum least squares twin support vector
  machine for classification,'' \emph{ACM Transactions on Internet Technology
  (TOIT)}, vol.~21, no.~3, pp. 1--24, 2021.

\bibitem[Richhariya and Tanveer(2020)]{richhariya2020reduced}
B.~Richhariya and M.~Tanveer, ``A reduced universum twin support vector machine
  for class imbalance learning,'' \emph{Pattern Recognition}, vol. 102, p.
  107150, 2020.

\bibitem[Kumari et~al.(2023)Kumari, Ganaie, and
  Tanveer]{kumari2023intuitionistic}
A.~Kumari, M.~A. Ganaie, and M.~Tanveer, ``Intuitionistic fuzzy universum
  support vector machine,'' in \emph{Neural Information Processing: 29th
  International Conference, ICONIP 2022, Virtual Event, November 22--26, 2022,
  Proceedings, Part I}.\hskip 1em plus 0.5em minus 0.4em\relax Springer, 2023,
  pp. 236--247.

\bibitem[Tanveer et~al.(2022, doi: 10.1109/TCYB.2022.3165879)Tanveer, Ganaie,
  Bhattacharjee, and Lin]{tanveer2022intuitionistic}
M.~Tanveer, M.~A. Ganaie, A.~Bhattacharjee, and C.~T. Lin, ``Intuitionistic
  fuzzy weighted least squares twin {SVM}s,'' \emph{IEEE Transactions on
  Cybernetics}, 2022, doi: 10.1109/TCYB.2022.3165879.

\bibitem[Kumari et~al.(2024)Kumari, Tanveer, and Lin]{anuradha2024}
A.~Kumari, M.~Tanveer, and C.~T. Lin, ``Class probability and generalized bell
  fuzzy twin {SVM} for imbalanced data,'' \emph{IEEE Transactions on Fuzzy
  Systems}, 2024.

\bibitem[Derrac et~al.(2015)Derrac, Garcia, Sanchez, and
  Herrera]{derrac2015keel}
J.~Derrac, S.~Garcia, L.~Sanchez, and F.~Herrera, ``K{EEL} data-mining software
  tool: Data set repository, integration of algorithms and experimental
  analysis framework,'' \emph{J. Mult. Valued Logic Soft Comput}, vol.~17, p.
  255–287, 2015.

\bibitem[Dem{\v{s}}ar(2006)]{demvsar2006statistical}
J.~Dem{\v{s}}ar, ``Statistical comparisons of classifiers over multiple data
  sets,'' \emph{The Journal of Machine Learning Research}, vol.~7, pp. 1--30,
  2006.

\bibitem[Iman and Davenport(1980)]{iman1980approximations}
R.~L. Iman and J.~M. Davenport, ``Approximations of the critical region of the
  fbietkan statistic,'' \emph{Communications in Statistics-Theory and Methods},
  vol.~9, no.~6, pp. 571--595, 1980.

\bibitem[Tanveer et~al.(2024,
  https://doi.org/10.1038/s44220-024-00237-x)Tanveer, Goel, Sharma, Malik,
  Beheshti, Del~Ser, Suganthan, and Lin]{tanveer2024ensemble}
M.~Tanveer, T.~Goel, R.~Sharma, A.~K. Malik, I.~Beheshti, J.~Del~Ser, P.~N.
  Suganthan, and C.~T. Lin, ``Ensemble deep learning for {A}lzheimer’s
  disease characterization and estimation,'' \emph{Nature Mental Health}, pp.
  1--13, 2024, https://doi.org/10.1038/s44220-024-00237-x.

\bibitem[Zadeh(1965)]{zadeh1965fuzzy}
L.~A. Zadeh, ``Fuzzy sets,'' \emph{Information and control}, vol.~8, no.~3, pp.
  338--353, 1965.

\bibitem[Ha et~al.(2013)Ha, Wang, and Chen]{ha2013support}
M.~Ha, C.~Wang, and J.~Chen, ``The support vector machine based on
  intuitionistic fuzzy number and kernel function,'' \emph{Soft Computing},
  vol.~17, pp. 635--641, 2013.

\end{thebibliography}

\clearpage
\section*{Supplementary Material}
\renewcommand{\thesection}{S.I}
\section{Intuitionistic Fuzzy Membership Scheme} 
The concept of the fuzzy set was introduced by Zadeh \cite{zadeh1965fuzzy} in 1965, whereas the intuitionistic fuzzy set (IFS) was put forth by Atanassov and Atanassov \cite{atanassov1999intuitionistic} in order to address challenges related to uncertainty, and it allows for a precise simulation of the situation using current information and observations \cite{ha2013support, quadir2024intuitionistic}. Here, the intuitionistic fuzzy membership scheme for the set $X$ at the kernel space is discussed by projecting the dataset $X$ onto a higher-dimensional space.

\begin{enumerate}

\item \textbf{Membership function}: 
The importance of each data point is evaluated through the membership function, which takes into account the distance between the sample and the center of its corresponding class in the kernel space. 
Samples positioned closer to the class center receive greater weighting compared to those located further away from the center. The membership function of each training sample $x_i$ is defined as follows:
\begin{equation}
    \vartheta(x_{i}) = 
            \begin{cases}
                  1 - \frac{\|\psi(x_{i}) - C^+\|}{r^+ + \hspace{0.2cm} \eta}, & y_{i} = +1\\ \\
                   1 - \frac{\|\psi(x_{i}) - C^-\|}{r^- + \hspace{0.2cm} \eta}, & y_{i} = -1
            \end{cases}
\end{equation}
Here, $\eta > 0 $ is a parameter, The center and radius of the negative (positive) class are denoted by $C-$ ($C+$) and $r-$ ($r+$), respectively. The function $\psi$ is a mapping that nonlinearly transforms data samples from a lower-dimensional space to a higher-dimensional space. \\
The calculation of the class center is performed in the following manner:
\begin{equation}
    C^+ = \frac{1}{m_{1}} \sum _{y_{i}= +1} \psi(x_{i}) \hspace{0.2cm} \text{and} \hspace{0.2cm} C^- = \frac{1}{m_{2}} \sum _{y_{i}= -1} \psi(x_{i}), 
\end{equation}
here $m_{1}$ ($m_{2}$) denotes the number of samples in the positive (negative ) class.

The radii of each class are determined as follows:
\begin{equation} r^{+}=\underset{y_{i}= +1}\max \Vert \ \psi(x_{i})-C^{+}\Vert \hspace{0.2cm} \text{and} \hspace{0.2cm} r^{-}=\underset{y_{i}= -1}\max \Vert \ \psi(x_{i})-C^{-}\Vert.
\end{equation}

\item \textbf{Non-membership Function:}
The non-membership function is defined as the ratio of the number of heterogeneous points to the total number of training samples in that neighbourhood $\mu(x_i)$. Therefore, the nonmembership function is defined as:

\begin{equation}
    \sigma(x_{i}) = (1 - \vartheta (x_{i})) \mu (x_{i})
\end{equation}

where $0 \leq \vartheta(x_{i}) + \sigma(x_{i}) \leq 1$ and the value of $\mu (x_{i})$ is determined as

\begin{equation} 
\mu (x_{i})=\frac{|\lbrace \psi(x_{j})|\Vert  \psi(x_{i})- \psi(x_{j})\Vert \leq \alpha,\,y_{j}\ne y_{i}\rbrace |}{|\lbrace x_{j}|\Vert  \psi(x_{i})- \psi(x_{j})\Vert \leq \eta \rbrace |} 
\end{equation}
where, the adjustable parameter is denoted by $\eta$ and $\lvert . \lvert$ indicates the cardinality of a set.

\item \textbf{Score Function:}
The degree of membership and nonmembership are calculated for each sample and then assigned an IFN to the training samples.  
As a result, the training set is denoted as: $X^* = \{(x_1,y_1,\vartheta_1,\sigma_1), \ldots, (x_n,y_n,\vartheta_n,\sigma_n)\} $, where $\vartheta_i$ and $\sigma_i$ denotes the degree of membership and non-membership of sample $x_i$, respectively. Score function is calculated as:
\begin{equation}
\label{eq:106}
s_{i}=\left\lbrace \begin{array}{lr} \vartheta _{i}, & \sigma _{i}=0,\\ 0, & \vartheta _{i}\leq \sigma _{i}, \\ \frac{1-\sigma _{i}}{2-\vartheta _{i}-\sigma _{i}}, & \text{others}. \end{array}\right.  
\end{equation}

\end{enumerate}
\renewcommand{\thesection}{S.II}
\section{Algorithm and Computational Complexity Analysis of the proposed IFUTSVM-ID model}
\label{Computational Complexity}
Let $u$, $m_1$, and $m_2$ denote the number of samples within the universum, positive (minority), and negative (majority) classes, respectively. We evaluate complexity based on the sample count, a widely adopted approach by assuming that $n$ is fixed. $\mathcal{O}(m_1^3+m_2^3)$ is the time complexity of TSVM \cite{khemchandani2007twin}. The inclusion of universum data samples elevates the time complexity, resulting in an overall time complexity for UTSVM \cite{qi2012twin} of $\mathcal{O}(m_1^3+m_2^3+u^3)$. The computation of the degree of membership in the proposed IFUTSVM-ID model involves several steps: calculating the class radius, computing the class center, and measuring the distance of each sample from the class center. Therefore, the complexity for determining the membership degree is $\mathcal{O}(1)+\mathcal{O}(1)+\mathcal{O}(m_1)+\mathcal{O}(m_2)$. For measuring the degree of non-membership, the computational complexity is $\mathcal{O}(m_1)+\mathcal{O}(m_2)$. Hence, the proposed IFUTSVM-ID model utilizes $\mathcal{O}(m_1)+\mathcal{O}(m_2)$ operations for assigning the score values. \\
The proposed IFUTSVM-ID requires solving the QPP in the dimension $m_1+m_2+u$. Therefore, $\mathcal{O}((m_1+m_2+u)^3)=\mathcal{O}(m_1^3+m_2^3+u^3)$ is the computational complexity. Let $IR > 1$ and $IR = \frac{m_2}{m_1}$, the time complexity can be equivalently formulated as: $\mathcal{O}(m_1^3+IR^3m_1^3+u^3) =\mathcal{O}(m_1^3(1+IR^3)+u^3)=\mathcal{O}(IR^3m_1^3+u^3)$. Therefore, $m_1 \ll m_2$, $\mathcal{O}(IR^3m_1^3+u^3) \ll \mathcal{O}(m_1^3+m_2^3+u^3)$. Hence, $\mathcal{O}(IR^3m_1^3+u^3) + \mathcal{O}(m_1)+\mathcal{O}(m_2)$ is the overall computational complexity of the proposed IFUTSVM-ID model. Therefore, the time complexity of the proposed IFUTSVM-ID model is lower than UTSVM. The algorithm of the proposed IFUTSVM-ID model is briefly described in the Algorithm \ref{IFUTSVM-ID classifier}.

\begin{algorithm}
\caption{IFUTSVM-ID}
\label{IFUTSVM-ID classifier}
\textbf{Input:} Traning sample $X_1 \in \mathbb{R}^{m_1 \times n}$, $X_2 \in \mathbb{R}^{m_2 \times n}$ and $U \in \mathbb{R}^{u \times n}$, where $u=m_2 - m_2$ and $g= \lceil \frac{m_{1}}{2} \rceil$.\\
\textbf{Output:} The weight vectors and bias for each class.\\
\vspace{-0.5cm}
\begin{algorithmic}[1]
\State Construct $X_2^* \in \mathbb{R}^{m_1 \times n}$ and $U^* \in \mathbb{R}^{g \times n}$ using randomly selected reduced data samples from $X_2$ and $U$, respectively. 
\State Compute score values $S_i$ for samples belonging to class $X_i$ using \eqref{eq:106}, for $i = 1, 2$ respectively.
\State To obtain the optimal hyperplanes for each class, solve equations (26) and (30) for the linear case, or equations (36) and (37) for the nonlinear case of the main file.
\State Testing sample is classified into class $+1$ or $-1$ using (31) or (38) for linear and non-linear case of the main file, respectively.
\end{algorithmic}
\end{algorithm}

\renewcommand{\thesection}{S.III}
\section{Experimental Setup and Experimental Results}
 In this section, we give a detailed description of the experimental setup and the experimental results on KEEL datasets with label noise for each model including baseline and
 proposed.
\subsection{Experimental Setup}
The experimental hardware environment comprises a PC equipped with an Intel(R) Xeon(R) Gold 6226R CPU @ $2.90$GHz and $128$ GB RAM, operating on Windows 11 possessing MATLAB R2023b. All datasets are randomly partitioned into training and testing subsets, with a ratio of $70:30$, respectively. The Gaussian kernel function is utilized in all experiments to map the input samples into a higher-dimensional space. The Gaussian kernel is defined as follows: $\mathscr{K}(p,q) = e^{-\frac{1}{2\mu^2}\|p-q\|^2},$ where $p, q \in \mathbb{R}^n,$ and $\mu$ is a kernel parameter. We employ a five-fold cross-validation and grid search approach to optimize the models' hyperparameters from the following ranges: $c_i = c_u = \{10^{-5}, 10^{-4},\ldots, 10^{5}\}$, for $i=1,2,3,4$, $\epsilon = \{0.1, 0.3, 0.5, 0.6\}$, and $\mu = \{2^{-5}, 2^{-4},\ldots, 2^{5}\}$. We set $c_1=c_2$ and $c_3=c_4$, to reduce the computational cost of the model. The random averaging method is used to generate universum data, which involves randomly selecting the same number of samples from each class and averaging them to obtain the universum data. For the baseline IFW-LSTSVM model, the hyperparameter $r$ is selected from the range $\{0.5, 1, 1.5, 2, 2.5\}$ and for the CGFTSVM-ID model, the hyperparameter $r$ is selected from the range $\{0.5, 0.625, 0.75, 0.875, 1\}$.

The generalization performance of the proposed IFUTSVM-ID models has been assessed by comparing them with baseline models across various metrics including $accuracy$ ($ACC$), $sensitivity$, $precision$, and $specificity$ $rates$. Mathematically,
\begin{align}
    Accuracy \hspace{0.1cm}(ACC) = \frac{\mathcal{TN}+\mathcal{TP}}{\mathcal{TN}+\mathcal{TP}+\mathcal{FP}+\mathcal{FN}},
\end{align}
 \begin{align}
     Sensitivity = \frac{\mathcal{TP}}{\mathcal{TP}+ \mathcal{FN}},
 \end{align}
 \begin{align}
     Precision = \frac{\mathcal{TP}}{\mathcal{FP}+\mathcal{TP}},
 \end{align}
\begin{align}
    Specificity = \frac{\mathcal{TN}}{\mathcal{FP}+\mathcal{TN}},
\end{align}
where ($\mathcal{TP}$) represents true positive, ($\mathcal{FN}$) represents the false negative, ($\mathcal{FP}$) represents the false positive, and ($\mathcal{TN}$) represents the true negative, respectively.

\subsection{Evaluation on KEEL Datasets with Added Label Noise}
The evaluation conducted using KEEL datasets reflects real-world scenarios. However, it's crucial to acknowledge that data impurities or noise may stem from various factors. To showcase the effectiveness of the proposed IFUTSVM-ID  model, especially under challenging conditions, we deliberately introduced label noise to certain datasets. We selected five datasets, namely ecoli-0-1\_vs\_5, aus, abalone9-18, checkerboard\_Data, and brwisconsin to assess the robustness of the models. To ensure fairness in model evaluation, we intentionally selected two datasets where the proposed IFUTSVM-ID model did not achieve the highest performance and three datasets where they achieved comparable results to an existing model with varying levels of label noise. For a comprehensive analysis, we introduced label noise at various levels, including $5\%$, $10\%$, $15\%$, and $20\%$, intentionally corrupting the labels of these datasets. Table \ref{UCI and KEEL results with label noise} displays the accuracies of all models for the selected datasets with $5\%$, $10\%$, $15\%$, and $20\%$ noise. Consistently, the proposed IFUTSVM-ID model demonstrates superior performance over baseline models, exhibiting higher ACC. Significantly, they maintain this leading performance despite the presence of noise. The average ACC of the proposed IFUTSVM-ID on the abalone9-18 dataset at various noise levels is $76.83\%$, surpassing the performance of the baseline models. On the checkerboard\_Data dataset and ecoli-0-1\_vs\_5, the average ACC of the proposed model at different noise levels is $84.18$ and $78.47\%$, respectively. On the aus dataset, the proposed model achieves an average ACC of $85.05\%$, which is slightly lower than the average ACC of the IF-USVM model. However, the ACC of the proposed IFUTSVM-ID model reaches $86.47\%$ at a $0\%$ level of label noise. At each noise level, IFUTSVM-ID emerges as the top performer, with an overall average ACC of $79.79\%$. By subjecting the model to rigorous conditions, we aim to showcase the exceptional performance and superiority of the proposed IFUTSVM-ID model, especially in challenging scenarios. The above findings underscore the significance of the proposed IFUTSVM-ID model as resilient solutions, capable of performing well in demanding conditions characterized by noise and impurities.

\renewcommand{\thetable}{S.I}
\begin{table*}[htp]
\centering
    \caption{Performance comparison of the proposed IFUTSVM-ID with the baseline models on KEEL datasets with label noise.}
    \label{UCI and KEEL results with label noise}
    \resizebox{1.00\linewidth}{!}{
\begin{tabular}{lccccccc}
\hline
Dataset & Noise & UTSVM \cite{qi2012twin} & RUTSVM-CIL \cite{richhariya2020reduced} & IF-USVM \cite{kumari2023intuitionistic} & IFW-LSTSVM \cite{tanveer2022intuitionistic} & CGFTSVM-ID  \cite{anuradha2024}& IFUTSVM-ID \\
& & $ACC (\%)$   &  $ACC (\%)$   &  $ACC (\%)$   &  $ ACC (\%)$  &  $ACC (\%)$  &  $ACC (\%)$   \\
 &  & $(C, C_u, \epsilon, \sigma)$ & $(C, C_u, \epsilon, \sigma)$ & $(C, C_u, \epsilon, \sigma)$ & $(C_1, C_2, r, \sigma)$ & $(C_1, C_2, r, \sigma)$ & $(C, C_u, \epsilon, \sigma)$ \\ \hline
abalone9-18 & $5\%$ & $76.8$ & $64.84$ & $75.39$ & $75.89$ & $75.89$ & $77.56$ \\
 &  & $(0.1, 1, 0.3, 1)$ & $(1, 1000, 0.3, 1)$ & $(1000, 10000, 0.2, 0.25)$ & $(0.001, 0.00001, 0.5, 32)$ & $(0.01, 0.1, 1, 4)$ & $(0.0001, 0.01, 0.3, 8)$ \\
 & $10\%$ & $72.69$ & $64.84$ & $59.36$ & $74.89$ & $75.26$ & $75.97$ \\
 &  & $(0.01, 0.1, 0.1, 32)$ & $(1, 1, 0.6, 32)$ & $(100, 1, 0.3, 0.125)$ & $(10000, 100000, 0.5, 4)$ & $(0.001, 0.00001, 0.5, 32)$ & $(0.0001, 0.1, 0.3, 2)$ \\
 & $15\%$ & $79.5$ & $70.32$ & $75.39$ & $75.89$ & $75.43$ & $73.01$ \\
 &  & $(0.01, 1, 0.3, 32)$ & $(0.1, 0.001, 0.6, 32)$ & $(1000, 10000, 0.2, 0.25)$ & $(10000, 100000, 0.5, 4)$ & $(0.1, 0.1, 0.5, 8)$ & $(10, 0.1, 0.6, 16)$ \\
 & $20\%$ & $76.35$ & $77.17$ & $79$ & $79.37$ & $79.04$ & $80.78$ \\
 &  & $(1, 100, 0.6, 16)$ & $(100, 0.00001, 0.6, 16)$ & $(10000, 1, 0.2, 0.125)$ & $(0.00001, 0.00001, 0.5, 0.03125)$ & $(0.0001, 1, 1, 8)$ & $(0.0001, 0.01, 0.3, 8)$ \\ \hline
Average ACC &  & $76.34$ & $69.29$ & $72.29$ & $76.51$ & $76.41$ & $76.83$ \\ \hline
aus & $5\%$ & $84.37$ & $85.51$ & $85.99$ & $85.51$ & $85.47$ & $86.98$ \\
 &  & $(0.001, 10, 0.5, 32)$ & $(0.00001, 0.001, 0.1, 16)$ & $(100000, 0.1, 0.1, 0.03125)$ & $(0.01, 0.01, 2, 4)$ & $(1000, 100, 0.5, 0.5)$ & $(10, 0.001, 0.6, 16)$ \\
 & $10\%$ & $85.41$ & $85.51$ & $85.99$ & $69.08$ & $84.37$ & $86.47$ \\
 &  & $(0.0001, 0.00001, 0.3, 16)$ & $(0.00001, 1, 0.1, 16)$ & $(1000, 0.01, 0.4, 0.03125)$ & $(1, 0.00001, 1, 32)$ & $(0.00001, 0.00001, 0.5, 1)$ & $(1, 0.1, 0.3, 32)$ \\
 & $15\%$ & $83.92$ & $67.15$ & $85.51$ & $65.7$ & $84.41$ & $84.54$ \\
 &  & $(0.01, 0.1, 0.5, 32)$ & $(0.00001, 0.001, 0.1, 16)$ & $(100000, 0.00001, 0.1, 0.0625)$ & $(0.00001, 0.00001, 0.5, 0.03125)$ & $(1000, 100, 0.5, 0.5)$ & $(1, 0.00001, 0.2, 1)$ \\
 & $20\%$ & $85.44$ & $59.42$ & $85.51$ & $82.61$ & $82.61$ & $82.19$ \\
 &  & $(0.001, 1000, 0.4, 32)$ & $(1, 10, 0.6, 16)$ & $(100000, 0.00001, 0.1, 0.0625)$ & $(0.00001, 0.001, 0.5, 0.03125)$ & $(0.00001, 0.00001, 0.5, 1)$ & $(1, 0.1, 0.5, 16)$ \\ \hline
Average ACC &  & $84.79$ & $74.4$ & $85.75$ & $75.73$ & $84.22$ & $85.05$ \\ \hline
brwisconsin & $5\%$ & $77.55$ & $65.69$ & $78.04$ & $77.55$ & $77.55$ & $75.69$ \\
 &  & $(0.0001, 1000, 0.5, 32)$ & $(100, 0.00001, 0.6, 16)$ & $(1000, 0.001, 0.1, 0.03125)$ & $(10000, 0.0001, 2, 16)$ & $(0.00001, 0.00001, 0.5, 1)$ & $(1000, 1, 0.1, 1)$ \\
 & $10\%$ & $76.57$ & $65.69$ & $77.06$ & $75.1$ & $76.57$ & $74$ \\
 &  & $(0.001, 0.1, 0.5, 32)$ & $(10, 0.01, 0.6, 16)$ & $(1000, 0.01, 0.1, 0.03125)$ & $(10, 0.001, 2, 16)$ & $(0.0001, 0.00001, 0.5, 16)$ & $(0.001, 0.0001, 0.3, 1)$ \\
 & $15\%$ & $72.65$ & $79.71$ & $75.59$ & $74.12$ & $72.16$ & $73.63$ \\
 &  & $(0.01, 0.001, 0.1, 32)$ & $(0.1, 0.01, 0.4, 1)$ & $(1, 100000, 0.3, 1)$ & $(0.00001, 0.001, 0.5, 4)$ & $(0.1, 1, 0.875, 8)$ & $(1, 1, 0.5, 4)$ \\
 & $20\%$ & $74.12$ & $76.57$ & $76.57$ & $75.1$ & $74.61$ & $74.31$ \\
 &  & $(0.0001, 1000, 0.5, 32)$ & $(100, 0.1, 0.6, 16)$ & $(1000, 0.01, 0.1, 0.03125)$ & $(0.00001, 0.0001, 0.5, 4)$ & $(0.001, 10, 1, 32)$ & $(100, 0.00001, 0.5, 8)$ \\ \hline
Average ACC &  & $75.22$ & $71.92$ & $76.82$ & $75.47$ & $75.22$ & $74.41$ \\ \hline
checkerboard\_Data & $5\%$ & $82.37$ & $81.51$ & $81.99$ & $82.51$ & $85.47$ & $85.51$ \\
 &  & $(1, 0.01, 0.4, 2)$ & $(0.00001, 0.001, 0.1, 16)$ & $(10, 10, 0.1, 1)$ & $(0.00001, 0.00001, 0.5, 0.03125)$ & $(0.0001, 1, 0.5, 16)$ & $(0.01, 0.01, 0.1, 2)$ \\
 & $10\%$ & $85.41$ & $85.51$ & $85.99$ & $69.08$ & $86.37$ & $86.47$ \\
 &  & $(1, 0.00001, 0.4, 2)$ & $(0.00001, 0.001, 0.1, 1)$ & $(0.00001, 0.00001, 0.5, 2)$ & $(0.00001, 0.0001, 0.5, 4)$ & $(1, 10, 0.625, 4)$ & $(0.1, 1, 0.1, 0.25)$ \\
 & $15\%$ & $82.92$ & $67.15$ & $80.51$ & $65.7$ & $81.41$ & $84.54$ \\
 &  & $(0.01, 0.1, 0.5, 32)$ & $(0.01, 1, 0.1, 1)$ & $(0.00001, 0.00001, 0.5, 2)$ & $(0.01, 0.1, 0.5, 4)$ & $(0.001, 1, 0.625, 4)$ & $(1, 10, 0.1, 1)$ \\
 & $20\%$ & $80.44$ & $59.42$ & $80.51$ & $79.61$ & $80.61$ & $80.19$ \\
 &  & $(0.0001, 1, 0.5, 16)$ & $(0.001, 1, 0.4, 1)$ & $(10, 0.00001, 0.1, 1)$ & $(0.001, 10, 2, 8)$ & $(0.1, 1, 0.625, 4)$ & $(1, 0.1, 0.5, 1)$ \\ \hline
Average ACC &  & $82.79$ & $73.4$ & $82.25$ & $74.23$ & $83.47$ & $84.18$ \\ \hline
ecoli-0-1\_vs\_5 & $5\%$ & $73.06$ & $79.17$ & $78.89$ & $73.06$ & $75.83$ & $79.17$ \\
 &  & $(1, 0.01, 0.2, 32)$ & $(10, 0.0001, 0.1, 32)$ & $(10000, 0.01, 0.1, 0.25)$ & $(0.1, 1, 1, 2)$ & $(0.00001, 1, 0.625, 16)$ & $(0.01, 0.01, 0.1, 2)$ \\
 & $10\%$ & $74.44$ & $79.17$ & $75.99$ & $73.06$ & $75.83$ & $77.78$ \\
 &  & $(0.0001, 0.01, 0.4, 0.125)$ & $(0.00001, 0.001, 0.1, 16)$ & $(100000, 0.01, 0.1, 0.125)$ & $(0.00001, 10, 0.5, 0.5)$ & $(0.00001, 0.01, 1, 8)$ & $(1000, 0.00001, 0.6, 8)$ \\
 & $15\%$ & $73.06$ & $78.89$ & $68.06$ & $73.06$ & $74.44$ & $72.22$ \\
 &  & $(1, 0.00001, 0.4, 2)$ & $(0.01, 1, 0.1, 1)$ & $(1, 1, 0.1, 1)$ & $(1, 0.01, 0.5, 4)$ & $(0.1, 1, 0.625, 4)$ & $(0.001, 1, 0.5, 32)$ \\
 & $20\%$ & $88.89$ & $87.22$ & $87.22$ & $83.06$ & $84.44$ & $84.72$ \\
 &  & $(0.00001, 0.001, 0.1, 32)$ & $(0.001, 1, 0.4, 1)$ & $(1000, 0.00001, 0.1, 0.03125)$ & $(100, 0.1, 2, 32)$ & $(0.00001, 1, 0.625, 16)$ & $(0.1, 1, 0.3, 1)$ \\ \hline
Average ACC &  & $77.36$ & $81.11$ & $77.54$ & $75.56$ & $77.64$ & $78.47$ \\ \hline
Overall Average ACC &  & $79.3$ & $74.02$ & $78.93$ & $75.5$ & $79.39$ & $79.79$ \\ \hline
\end{tabular}}
\end{table*}

\renewcommand{\thetable}{S.II}
\begin{table*}[]
\centering
    \caption{Comparison of the performance of the proposed IFUTSVM-ID model with the baseline models based on classification ACC on KEEL datasets.}
    \label{Classification performance on KEEL dataset.}
    \resizebox{1\textwidth}{!}{
\begin{tabular}{lcccccc}
\hline
Dataset & UTSVM \cite{qi2012twin}& RUTSVM-CIL \cite{richhariya2020reduced}&  IF-USVM \cite{kumari2023intuitionistic}& IFW-LSTSVM \cite{tanveer2022intuitionistic}& CGFTSVM-ID \cite{anuradha2024}& IFUTSVM-ID \\
& $ACC (\%)$   &  $ACC (\%)$   &  $ACC (\%)$   &  $ ACC (\%)$  &  $ACC (\%)$  &  $ACC (\%)$   \\
 & $(C, C_u, \epsilon, \sigma)$ & $(C, C_u, \epsilon, \sigma)$ & $(C, C_u, \epsilon, \sigma)$ & $(C_1, C_2, r, \sigma)$ & $(C_1, C_2, r, \sigma)$ & $(C, C_u, \epsilon, \sigma)$ \\ \hline
abalone9-18 & $82.38$ & $88.73$ & $72.06$ & $89.89$ & $89.8$ & $89.95$ \\
$(731\times 7)$ & $(0.1, 1, 0.3, 1)$ & $(1, 1000, 0.3, 1)$ & $(10000, 10000, 0.6, 1)$ & $(100, 100, 0.5, 8)$ & $(0.0001, 1, 1, 8)$ & $(0.1, 1, 0.1, 1)$ \\
aus & $80.54$ & $80.71$ & $80.16$ & $81.16$ & $86.34$ & $86.47$ \\
$(690\times 14)$ & $(0.01, 0.00001, 0.1, 32)$ & $(0.00001, 1, 0.1, 2)$ & $(1000, 10000, 0.2, 0.25)$ & $(10000, 100000, 0.5, 4)$ & $(0.1, 0.1, 0.5, 8)$ & $(0.0001, 0.1, 0.3, 2)$ \\
brwisconsin & $96.4$ & $65.69$ & $96.74$ & $86.76$ & $96.57$ & $100$ \\
$(683\times 9)$ & $(1, 100, 0.6, 16)$ & $(0.00001, 1, 0.1, 16)$ & $(100, 1, 0.3, 0.125)$ & $(0.001, 0.00001, 0.5, 32)$ & $(0.001, 0.00001, 0.5, 32)$ & $(0.1, 0.00001, 0.3, 16)$ \\
checkerboard\_Data & $90.54$ & $87.71$ & $89.16$ & $81.16$ & $90.34$ & $86.47$ \\
$(690\times 14)$ & $(0.01, 0.1, 0.1, 32)$ & $(0.00001, 100, 0.1, 2)$ & $(1000, 10000, 0.2, 0.25)$ & $(10000, 100000, 0.5, 4)$ & $(0.1, 0.1, 0.5, 8)$ & $(0.0001, 0.1, 0.3, 2)$ \\
bupa or liver-disorders & $63.1$ & $66.22$ & $69.07$ & $53.4$ & $69.9$ & $70.51$ \\
$(345\times 6)$ & $(0.01, 1000, 0.4, 4)$ & $(0.1, 100, 0.5, 8)$ & $(10000, 1, 0.2, 0.125)$ & $(10000, 0.0001, 2, 16)$ & $(0.01, 0.1, 1, 4)$ & $(0.0001, 0.01, 0.3, 8)$ \\
ecoli-0-1\_vs\_5 & $80$ & $76.57$ & $79.25$ & $93.06$ & $97.22$ & $97.22$ \\
$(240\times 6)$ & $(0.01, 0.001, 0.1, 32)$ & $(1, 1, 0.6, 32)$ & $(100000, 0.01, 0.4, 0.0625)$ & $(100, 100, 2, 2)$ & $(0.00001, 0.01, 0.625, 8)$ & $(1000, 0.01, 0.6, 16)$ \\
ecoli-0-1\_vs\_2-3-5 & $87.5$ & $78.8$ & $83.15$ & $85$ & $86.89$ & $87.34$ \\
$(244\times 7)$ & $(0.0001, 0.00001, 0.3, 16)$ & $(100, 1, 0.4, 32)$ & $(1000, 0.001, 0.1, 0.03125)$ & $(1, 0.001, 0.5, 32)$ & $(0.00001, 0.001, 0.5, 1)$ & $(10, 0.1, 0.6, 16)$ \\
ecoli-0-1-4-6\_vs\_5 & $75$ & $89.74$ & $81.67$ & $90$ & $90$ & $91.67$ \\
$(280\times 6)$ & $(0.01, 0.00001, 0.3, 32)$ & $(0.1, 0.001, 0.6, 32)$ & $(100000, 0.1, 0.1, 0.03125)$ & $(10, 1000, 2, 8)$ & $(0.00001, 0.01, 0.625, 8)$ & $(10, 0.01, 0.6, 32)$ \\
ecoli0137vs26 & $92.5$ & $92.5$ & $92.5$ & $70.97$ & $94.34$ & $94.62$ \\
$(311\times 7)$ & $(0.01, 1, 0.3, 32)$ & $(10, 0.01, 0.6, 16)$ & $(100000, 0.01, 0.4, 0.0625)$ & $(0.00001, 0.00001, 0.5, 0.03125)$ & $(0.001, 0.001, 0.5, 1)$ & $(0.01, 1, 0.1, 1)$ \\
ecoli01vs5 & $80$ & $80$ & $70$ & $70$ & $80$ & $80$ \\
$(240\times 7)$ & $(0.01, 0.1, 0.5, 32)$ & $(10, 0.0001, 0.1, 32)$ & $(100000, 0.00001, 0.1, 0.0625)$ & $(1, 0.00001, 1, 32)$ & $(0.00001, 0.00001, 0.5, 1)$ & $(0.00001, 0.01, 0.1, 1)$ \\
ecoli-0-1-4-7\_vs\_2-3-5-6 & $75$ & $80.56$ & $79.44$ & $85.92$ & $84.89$ & $88.33$ \\
$(336\times 7)$ & $(0.001, 10, 0.5, 32)$ & $(100000, 0.0001, 0.6, 8)$ & $(100000, 0.01, 0.5, 0.0625)$ & $(0.01, 0.01, 2, 4)$ & $(0.00001, 0.001, 0.5, 1)$ & $(10, 0.0001, 0.4, 32)$ \\
ecoli-0-1-4-7\_vs\_5-6& $78.57$ & $80$ & $76.94$ & $0$ & $81.86$ & $81.91$ \\
$(332\times 6)$ & $(0.0001, 1, 0.5, 16)$ & $(1, 1000, 0.6, 16)$ & $(1000, 0.01, 0.4, 0.03125)$ & $(1, 1, 2, 8)$ & $(0.001, 10, 0.625, 8)$ & $(10, 0.001, 0.6, 16)$ \\
ecoli-0-3-4-6\_vs\_5 & $83.33$ & $82.47$ & $83.33$ & $80.78$ & $82.56$ & $85.47$ \\
$(205\times 7)$ & $(0.001, 1000, 0.4, 32)$ & $(1, 10, 0.6, 16)$ & $(100000, 0.00001, 0.1, 0.0625)$ & $(0.00001, 0.00001, 0.5, 0.03125)$ & $(0.00001, 0.00001, 0.5, 1)$ & $(1, 0.1, 0.3, 32)$ \\
ecoli-0-4-6\_vs\_5 & $83.33$ & $90.74$ & $91.67$ & $90$ & $96.67$ & $98.33$ \\
$(203\times 6)$ & $(0.1, 10, 0.3, 32)$ & $(100, 0.1, 0.6, 16)$ & $(1000, 0.01, 0.1, 0.03125)$ & $(0.00001, 0.001, 0.5, 0.03125)$ & $(1000, 100, 0.5, 0.5)$ & $(1, 0.1, 0.5, 16)$ \\
ecoli-0-3-4-7\_vs\_5-6 & $87.5$ & $78.8$ & $86.78$ & $83.15$ & $92.4$ & $92.55$ \\
$(257\times 7)$ & $(0.0001, 1000, 0.5, 32)$ & $(100, 0.00001, 0.6, 16)$ & $(1000, 0.001, 0.1, 0.03125)$ & $(0.00001, 0.001, 0.5, 0.03125)$ & $(1000, 100, 0.5, 0.5)$ & $(1000, 0.01, 0.6, 8)$ \\
ecoli3 & $87.78$ & $80.56$ & $72.78$ & $90$ & $90$ & $91$ \\
$(336\times 7)$ & $(0.00001, 0.00001, 0.1, 1)$ & $(0.00001, 0.001, 0.1, 1)$ & $(1, 0.1, 0.2, 1)$ & $(0.00001, 0.00001, 0.5, 0.0625)$ & $(0.0001, 0.00001, 0.5, 16)$ & $(1, 0.00001, 0.2, 1)$ \\
ecoli2 & $87.78$ & $80.56$ & $62.78$ & $90$ & $91$ & $91$ \\
$(336\times 7)$ & $(1, 100, 0.2, 2)$ & $(0.01, 0.00001, 0.1, 1)$ & $(1, 1, 0.1, 1)$ & $(0.00001, 0.001, 0.5, 4)$ & $(0.0001, 0.001, 0.5, 1)$ & $(1, 1, 0.2, 1)$ \\
ecoli4 & $93.75$ & $92.12$ & $93.75$ & $92$ & $91$ & $96$ \\
$(336\times 7)$ & $(1, 0.01, 0.4, 2)$ & $(0.1, 0.01, 0.4, 1)$ & $(1, 1, 0.4, 1)$ & $(1, 0.0001, 0.5, 32)$ & $(0.001, 0.0001, 0.625, 1)$ & $(0.01, 0.1, 0.1, 1)$ \\
ecoli-0-2-3-4\_vs\_5 & $81.48$ & $89.82$ & $81.48$ & $95.89$ & $97.89$ & $98.7$ \\
$(202\times 7)$ & $(0.001, 0.00001, 0.4, 32)$ & $(0.00001, 0.001, 0.1, 16)$ & $(1000, 0.00001, 0.1, 0.03125)$ & $(1000, 1, 0.5, 2)$ & $(1, 10, 0.5, 4)$ & $(100, 0.01, 0.6, 32)$ \\
ecoli-0-6-7\_vs\_3-5 & $83.33$ & $74.77$ & $83.86$ & $88.91$ & $89.45$ & $89.39$ \\
$(222\times 7)$ & $(0.001, 0.1, 0.5, 32)$ & $(10, 0.00001, 0.1, 16)$ & $(1000, 0.01, 0.3, 0.03125)$ & $(0.00001, 0.00001, 0.5, 0.03125)$ & $(1000, 100, 0.5, 0.5)$ & $(1, 0.1, 0.6, 16)$ \\
glass4 & $90.44$ & $90.31$ & $89.75$ & $87.31$ & $90.88$ & $91.98$ \\
$(214\times 9)$ & $(1, 0.00001, 0.4, 2)$ & $(0.1, 0.001, 0.4, 1)$ & $(1, 1, 0.4, 1)$ & $(100000, 10000, 1, 1)$ & $(1000, 100, 0.5, 4)$ & $(0.0001, 0.1, 0.4, 4)$ \\
glass5 & $96.88$ & $96.88$ & $95.31$ & $96.88$ & $96.88$ & $96.88$ \\
$(214\times 9)$ & $(1, 0.00001, 0.2, 8)$ & $(0.01, 1, 0.1, 1)$ & $(1, 100000, 0.3, 1)$ & $(10, 0.001, 2, 16)$ & $(10, 1, 0.625, 4)$ & $(1000, 1, 0.1, 1)$ \\
haber & $75.82$ & $75.02$ & $75.82$ & $75.82$ & $75.82$ & $75.82$ \\
$(306\times 3)$ & $(0.0001, 0.01, 0.4, 0.125)$ & $(0.001, 1, 0.4, 1)$ & $(1, 0.1, 0.1, 2)$ & $(100000, 0.01, 1, 16)$ & $(100, 10, 0.75, 4)$ & $(0.0001, 1, 0.4, 4)$ \\
heart-stat & $72.84$ & $56.79$ & $56.79$ & $75.31$ & $82.42$ & $84.44$ \\
$(270\times 13)$ & $(0.001, 10, 0.3, 2)$ & $(0.0001, 0.00001, 0.2, 4)$ & $(100, 1, 0.1, 0.5)$ & $(1, 0.00001, 1, 16)$ & $(1, 1, 0.5, 4)$ & $(10, 0.1, 0.4, 16)$ \\
iono & $93.33$ & $65.71$ & $90.48$ & $69.52$ & $92.38$ & $95.71$ \\
$(351\times 33)$ & $(0.00001, 10000, 0.1, 0.5)$ & $(10, 100, 0.2, 1)$ & $(100, 100, 0.1, 1)$ & $(1, 10, 2, 8)$ & $(0.1, 1, 0.875, 8)$ & $(0.00001, 1, 0.1, 8)$ \\
led7digit-0-2-4-5-6-7-8-9\_vs\_1 & $95.46$ & $95.46$ & $91.67$ & $89.39$ & $94.97$ & $95.7$ \\
$(443\times 7)$ & $(1, 1, 0.2, 32)$ & $(0.001, 100000, 0.3, 4)$ & $(1000, 0.1, 0.1, 0.0625)$ & $(0.00001, 1000, 0.5, 2)$ & $(0.00001, 10, 0.5, 32)$ & $(0.1, 1, 0.3, 1)$ \\
monk1 & $43.37$ & $50.6$ & $43.98$ & $45.78$ & $45.78$ & $52.53$ \\
$(556\times 6)$ & $(1, 0.01, 0.2, 32)$ & $(10, 1000, 0.5, 32)$ & $(1000, 0.1, 0.1, 0.0625)$ & $(10000, 100, 0.5, 2)$ & $(0.00001, 0.00001, 0.5, 16)$ & $(0.00001, 0.1, 0.6, 0.25)$ \\
monk2 & $81.67$ & $65.56$ & $56.67$ & $67.22$ & $80.89$ & $81.67$ \\
$(601\times 7)$ & $(0.1, 1000, 0.1, 32)$ & $(100000, 0.01, 0.5, 2)$ & $(1000, 0.01, 0.1, 0.03125)$ & $(0.00001, 0.0001, 0.5, 4)$ & $(0.001, 10, 1, 32)$ & $(0.00001, 0.1, 0.1, 0.03125)$ \\
monk3 & $40.96$ & $64.46$ & $42.77$ & $45.18$ & $46.99$ & $66.51$ \\
$(554\times 6)$ & $(0.001, 0.0001, 0.4, 1)$ & $(0.00001, 0.001, 0.1, 4)$ & $(1000, 1, 0.4, 1)$ & $(0.01, 0.1, 0.5, 4)$ & $(0.00001, 0.01, 0.5, 4)$ & $(0.001, 0.0001, 0.3, 1)$ \\
new-thyroid1 & $98.44$ & $96.88$ & $95.31$ & $95.44$ & $93.85$ & $96.88$ \\
$(215\times 5)$ & $(0.01, 1, 0.1, 2)$ & $(0.1, 0.001, 0.1, 1)$ & $(10, 10, 0.1, 1)$ & $(1, 0.01, 2, 4)$ & $(10, 1, 0.5, 4)$ & $(0.01, 1, 0.6, 4)$ \\
pima & $69.57$ & $66.09$ & $73.04$ & $61.3$ & $70.26$ & $71.74$ \\
$(768\times 8)$ & $(100, 10, 0.3, 8)$ & $(0.01, 1, 0.3, 0.5)$ & $(0.00001, 0.00001, 0.5, 2)$ & $(0.00001, 0.00001, 0.5, 0.03125)$ & $(10, 1, 0.5, 4)$ & $(0.001, 1, 0.5, 32)$ \\
segment0 & $95.71$ & $96.24$ & $97.54$ & $85.4$ & $97.85$ & $98.12$ \\
$(2308\times 19)$ & $(0.00001, 0.1, 0.1, 0.03125)$ & $(0.001, 1, 0.5, 1)$ & $(10000, 1, 0.1, 0.5)$ & $(0.00001, 0.00001, 0.5, 0.03125)$ & $(0.0001, 1, 0.5, 16)$ & $(1, 1, 0.5, 4)$ \\
shuttle-c0-vs-c4 & $99.82$ & $99.82$ & $99.82$ & $94.16$ & $95$ & $99.82$ \\
$(1829\times 9)$ & $(1, 0.001, 0.5, 16)$ & $(0.001, 0.1, 0.5, 4)$ & $(100000, 1, 0.2, 0.25)$ & $(10, 100, 1, 2)$ & $(0.1, 1, 0.625, 4)$ & $(0.00001, 0.1, 0.1, 16)$ \\
shuttle-6\_vs\_2-3 & $98.55$ & $100$ & $98.55$ & $98.55$ & $100$ & $100$ \\
$(230\times 9)$ & $(0.00001, 0.001, 0.1, 32)$ & $(0.001, 1, 0.3, 0.5)$ & $(10000, 0.01, 0.1, 0.125)$ & $(0.001, 10, 2, 8)$ & $(0.1, 1, 0.5, 32)$ & $(1, 10, 0.4, 32)$ \\
transfusion & $73.21$ & $71.43$ & $61.16$ & $67.86$ & $76.79$ & $78.13$ \\
$(748\times 4)$ & $(0.00001, 10, 0.5, 16)$ & $(0.001, 1, 0.3, 32)$ & $(1000, 100000, 0.1, 0.03125)$ & $(1000, 10000, 2, 4)$ & $(0.001, 1, 0.625, 4)$ & $(100, 0.00001, 0.5, 8)$ \\
vehicle1 & $75.49$ & $71.94$ & $69.96$ & $76.68$ & $74.58$ & $69.96$ \\
$(846\times 18)$ & $(0.0001, 100000, 0.5, 8)$ & $(0.00001, 1, 0.6, 8)$ & $(10000, 1, 0.5, 0.125)$ & $(0.1, 1, 1, 2)$ & $(0.00001, 0.01, 1, 8)$ & $(100000, 0.0001, 0.6, 8)$ \\
vehicle2 & $96.05$ & $80.63$ & $79.05$ & $73.52$ & $97.23$ & $80.63$ \\
$(846\times 18)$ & $(0.00001, 1000, 0.1, 0.03125)$ & $(1, 10, 0.4, 32)$ & $(10, 10, 0.5, 0.03125)$ & $(1000, 1, 1, 8)$ & $(0.00001, 0.1, 1, 16)$ & $(1000, 0.00001, 0.6, 8)$ \\
votes & $90.92$ & $90$ & $87.69$ & $79.23$ & $92.15$ & $93.85$ \\
$(435\times 16)$ & $(0.00001, 0.001, 0.1, 16)$ & $(0.00001, 0.01, 0.1, 16)$ & $(1000, 0.00001, 0.1, 0.03125)$ & $(100, 0.1, 2, 32)$ & $(0.001, 0.01, 0.5, 2)$ & $(0.01, 0.01, 0.1, 2)$ \\
yeast-0-2-5-7-9\_vs\_3-6-8 & $93.01$ & $93.69$ & $93.36$ & $91.03$ & $95.01$ & $95.02$ \\
$(1004\times 8)$ & $(0.01, 0.01, 0.6, 32)$ & $(1, 0.1, 0.6, 16)$ & $(10000, 100, 0.3, 0.0625)$ & $(100, 100, 0.5, 8)$ & $(0.1, 1, 1, 4)$ & $(1, 0.1, 0.5, 1)$ \\
yeast-0-2-5-6\_vs\_3-7-8-9 & $91.68$ & $87.04$ & $91.69$ & $91.02$ & $90.36$ & $92.03$ \\
$(194\times 33)$ & $(0.0001, 0.00001, 0.1, 8)$ & $(0.001, 1, 0.5, 4)$ & $(100000, 0.01, 0.1, 0.125)$ & $(1, 0.01, 0.5, 4)$ & $(0.1, 1, 0.75, 8)$ & $(1, 10000, 0.2, 1)$ \\
yeast1 & $91.16$ & $74.03$ & $91.56$ & $71.91$ & $74.16$ & $92.14$ \\
$(1484\times 8)$ & $(0.1, 1, 0.5, 32)$ & $(1000, 0.01, 0.6, 8)$ & $(100000, 0.00001, 0.1, 0.0625)$ & $(100, 0.1, 2, 1)$ & $(1, 10, 0.625, 4)$ & $(1, 0.01, 0.3, 1)$ \\
yeast1vs7 & $74.38$ & $71.24$ & $73.03$ & $80.16$ & $80$ & $80.29$ \\
$(459\times 8)$ & $(10, 0.001, 0.2, 32)$ & $(0.001, 0.0001, 0.1, 2)$ & $(10000, 0.01, 0.1, 0.25)$ & $(0.00001, 10, 0.5, 0.5)$ & $(0.0001, 0.00001, 0.875, 4)$ & $(1, 10, 0.1, 1)$ \\
yeast2vs8 & $92.22$ & $96.53$ & $91.56$ & $95.83$ & $96.92$ & $97.22$ \\
$(483\times 8)$ & $(0.00001, 0.1, 0.1, 0.03125)$ & $(0.0001, 0.01, 0.1, 4)$ & $(100000, 1, 0.3, 0.0625)$ & $(0.00001, 0.00001, 0.5, 0.03125)$ & $(0.0001, 1, 0.625, 8)$ & $(0.01, 1, 0.3, 1)$ \\
yeast-2\_vs\_4 & $77.08$ & $75.56$ & $71.24$ & $71.56$ & $77.89$ & $79.22$ \\
$(514\times 8)$ & $(0.1, 0.001, 0.2, 1)$ & $(0.001, 0.01, 0.6, 1)$ & $(10, 0.00001, 0.1, 1)$ & $(0.00001, 10, 1, 0.5)$ & $(0.00001, 1, 0.625, 16)$ & $(10, 0.01, 0.2, 1)$ \\
yeast5 & $77.75$ & $80.64$ & $78.91$ & $80.87$ & $80.75$ & $81.35$ \\
$(1484\times 8)$ & $(0.1, 0.00001, 0.1, 0.5)$ & $(0.1, 0.1, 0.1, 1)$ & $(1, 1, 0.1, 1)$ & $(0.00001, 0.00001, 0.5, 0.03125)$ & $(0.00001, 1, 0.625, 16)$ & $(0.1, 1, 0.1, 0.25)$ \\
yeast3 & $86.18$ & $83.48$ & $84.92$ & $89.21$ & $85.61$ & $89.44$ \\
$(1484\times 8)$ & $(0.1, 0.001, 0.3, 0.25)$ & $(0.1, 0.01, 0.3, 1)$ & $(1, 1, 0.5, 1)$ & $(10000, 0.01, 1, 8)$ & $(0.0001, 1, 0.625, 8)$ & $(1, 1, 0.6, 1)$ \\
 \hline
Average ACC & $83.52$ & $81.15$ & $80.61$ & $79.44$ & $86.19$ & $87.70$ \\ \hline
Average Rank & $3.71$ & $4.24$ & $4.47$ & $4.33$ & $2.76$ & $1.54$ \\ \hline
\end{tabular}}
\end{table*}

\end{document}